\documentclass[11pt]{article}

\usepackage[margin=1in]{geometry}
\usepackage{amsmath,amssymb,bm}
\usepackage{graphicx}
\usepackage{subfig}
\usepackage{booktabs}
\usepackage{multirow}
\usepackage{makecell}
\usepackage[numbers,sort&compress]{natbib}
\usepackage{float}
\usepackage[colorlinks=true,linkcolor=blue,citecolor=blue,urlcolor=blue]{hyperref}
\usepackage{url}

\title{Spatiotemporal system forecasting with irregular time steps via masked autoencoder}

\author{
Kewei Zhu\textsuperscript{1,*}\quad
Yanze Xin\textsuperscript{2,*}\quad
Jinwei Hu\textsuperscript{3}\quad
Xiaoyuan Cheng\textsuperscript{4}\quad
Yiming Yang\textsuperscript{5,6}\quad
Sibo Cheng\textsuperscript{7,\textdagger}
}

\date{
\small
\textsuperscript{1}Department of Chemical Engineering, University College London, UK\\
\textsuperscript{2}Department of Computing, Imperial College London, UK\\
\textsuperscript{3}Department of Computer Science, University of Liverpool, UK\\
\textsuperscript{4}Dynamic Systems Lab, University College London, UK\\
\textsuperscript{5}Department of Civil, Environmental \& Geomatic Engineering, University College London, UK\\
\textsuperscript{6}Department of Statistical Science, University College London, UK\\
\textsuperscript{7}CEREA, ENPC, EDF R\&D, Institut Polytechnique de Paris, France\\[4pt]
\textsuperscript{*}These authors contributed equally to this work.\\
\textsuperscript{\textdagger}Corresponding author: \texttt{sibo.cheng@enpc.fr}
}

\begin{document}
\maketitle
\thispagestyle{plain}
\renewcommand{\thefootnote}{}
\footnotetext{This is the accepted manuscript of an article accepted for publication in \textit{Physica D: Nonlinear Phenomena}.}
\renewcommand{\thefootnote}{\arabic{footnote}}

\begin{abstract}
Predicting high-dimensional dynamical systems with irregular time steps presents significant challenges for current data-driven algorithms.
These irregularities arise from missing data, sparse observations, or adaptive computational techniques, reducing prediction accuracy.
To address these limitations, we propose a novel method: a Physics-Spatiotemporal Masked Autoencoder.
This method integrates convolutional autoencoders for spatial feature extraction with masked autoencoders optimised for irregular time series, leveraging attention mechanisms to reconstruct the entire physical sequence in a single prediction pass.
The model avoids the need for data imputation while preserving the physical integrity of the system.
Here, 'physics' refers to high-dimensional fields generated from underlying dynamical systems, rather than enforcing explicit physical constraints or PDE residuals.
We evaluate this approach on multiple simulated datasets and real-world ocean temperature data.
The results demonstrate that our method achieves significant improvements in prediction accuracy, robustness to nonlinearities, and computational efficiency over traditional convolutional and recurrent network methods. 
The model shows potential for capturing complex spatiotemporal patterns without requiring domain-specific knowledge, with applications in climate modelling, fluid dynamics, ocean forecasting, environmental monitoring, and scientific computing.
\end{abstract}


\section*{Acronyms}
\begin{tabular}{@{}ll@{}}
  P-STMAE & Physics Spatiotemporal Masked Autoencoder \\
  PDE & Partial Differential Equation \\
  ARIMA & Auto-Regressive Integrated Moving Average \\
  MLP & Multi-Layer Perceptron \\
  CNN & Convolutional Neural Network \\
  RNN & Recurrent Neural Network \\
  M-RNN & Multi-directional Recurrent Neural Network \\
  Seq2Seq & Sequence-to-Sequence \\
  GRU & Gated Recurrent Unit \\
  LSTM & Long Short-Term Memory Network \\
  ConvRAE & Deep Convolutional Recurrent Autoencoder \\
  ConvLSTM & Convolutional Long Short-Term Memory Network \\
  FC-LSTM & Fully Connected LSTM \\
  NLP & Natural Language Processing \\
  BERT & Bidirectional Encoder Representations from Transformers \\
  ROM & Reduced Order Modelling \\
  POD & Proper Orthogonal Decomposition \\
  DMD & Dynamic Mode Decomposition \\
  PCA & Principal Component Analysis \\
  CAE & Convolutional Autoencoder \\
  MAE & Masked Autoencoder \\
  TiMAE & Time Series Masked Autoencoder \\
  MSE & Mean Squared Error \\
  SSIM & Structural Similarity Index Measure \\
  PSNR & Peak Signal-to-Noise Ratio \\
  SWE & Shallow Water Equation \\
  SST & NOAA Sea Surface Temperature \\
  VAE & Variational Autoencoder \\
  GAN & Generative Adversarial Network \\
  MSPCNN & Multi-Scale Physics-Constrained Neural Network \\
  PINN & Physics-Informed Neural Network \\
  KAN & Kolmogorov-Arnold Network \\
\end{tabular}

\section*{Main Notations}
\begin{tabular}{@{}ll@{}}
  $\mathbf{x}_t$ & Observation state in the full space at time step $t$  \\
  $\hat{\mathbf{x}}_t$ & Predicted state in the full space at time step $t$  \\
  $\mathbf{z}_t$ & Observation state in the latent space at time step $t$  \\
  $\hat{\mathbf{z}}_t$ & Predicted state in the latent space at time step $t$  \\
  $T_{\text{in}}$ & Set of input time steps  \\
  $T_{\text{out}}$ & Set of forecasting time steps  \\
  $T_{\text{miss}}$ & Set of missing time steps of the input  \\
  $T_{\text{obs}}$ & Set of observed time steps of the input  \\
  $t_{\text{in}}$ & Number of input time steps  \\
  $t_{\text{out}}$ & Number of forecasting time steps  \\
  $\mathbf{X}_T$ & Physical states for the set of time steps $T$  \\
  $\hat{\mathbf{X}}$ & Reconstructed sequence in the full space  \\
  $\mathbf{Z}_T$ & Latent states for the set of time steps $T$  \\
  $\hat{\mathbf{Z}}$ & Reconstructed sequence in the latent space  \\
  $\Phi_x$ & Placeholders in the full space  \\
  $\phi_x$ & Physical placeholder state  \\
  $\Phi_z$ & Placeholders in the latent space  \\
  $\phi_z$ & Latent placeholding state  \\
  $d_x$ & Dimension of the physical state  \\
  $d_z$ & Dimension of the latent state  \\
  $d_k$ & Dimension of the attention vector  \\
  $\mathbf{Q},\,\mathbf{K},\,\mathbf{V}$ & Attention matrices of transformer blocks  \\
  $\mathbf{W}_Q,\,\mathbf{W}_K,\,\mathbf{W}_V$ & Trainable weight matrices of transformer blocks  \\
  $\mathbf{A}$ & Attention weights matrix  \\
  $\mathbf{O}$ & Output matrix from transformer block  \\
  $\mathbf{O'}$ & Final output matrix from transformer block  \\
  $f_E$ & Encoder function mapping physical to latent space  \\
  $f_D$ & Decoder function mapping latent to physical space  \\
  $\theta_E$ & Parameters of the encoder  \\
  $\theta_D$ & Parameters of the decoder  \\
  $\delta_t$ & Positional embedding at time step $t$  \\
  $\mathcal{L}$ & Overall loss function  \\
  $\lambda$ & Weighting coefficient for the latent loss \\
\end{tabular}

\section{Introduction}\label{Xsec1-1}
\label{ch:introduction}

Predicting high-dimensional dynamical systems is challenging when observations occur at irregular time intervals.
Such irregularities frequently arise in scientific simulations and observational datasets due to sensor failures, sparse measurement networks~\cite{fukami2021global}, or adaptive time stepping in numerical solvers based on Partial Differential Equations (PDEs)~\cite{caffarelli2010rates,brin2002introduction,brunton2017chaos}.
Conventional machine learning models, such as Multi-Layer Perceptrons (MLPs) and  Recurrent Neural Networks (RNNs), typically assume regularly sampled data and struggle to generalise when faced with temporal gaps or uneven sampling~\cite{siami2019performance}.
To compensate, many workflows rely on preprocessing techniques~\cite{weerakody2021review}, such as resampling, interpolation~\cite{johnson2022empirical}, or data assimilation~\cite{cheng2024efficient} to produce uniformly spaced sequences.
However, these procedures can introduce bias, increase computational cost, and obscure the true temporal dynamics of the system~\cite{afrifa2020missing,ahn2022comparison}.
There is a clear need for models that can directly learn from irregular time series without preprocessing while accurately capturing the underlying spatiotemporal structure of the physical system.

Traditional approaches to handling temporally irregular observations include time series models like Auto-Regressive Integrated Moving Average (ARIMA)~\cite{nelson1998time} and data assimilation algorithms such as Kalman Filters~\cite{gomez1994estimation}.
While ARIMA models are effective for univariate, stationary time series, they face limitations with high-dimensional, nonlinear systems~\cite{kontopoulou2023review,siami2018comparison}.
Kalman Filters estimate the state of a linear system from incomplete measurements but rely on assumptions of linearity and normality that often do not hold in complex, high-dimensional dynamics~\cite{carrassi2018data}.
Extended and Unscented Kalman Filters~\cite{ribeiro2004kalman,wan2000unscented,cheng2023machine,gleiter2022ensemble} attempt to address nonlinearities but still struggle with high dimensionality and irregular time steps common in real-world data.

\begin{figure*}[!t]

\centerline{\includegraphics[width=0.7\textwidth]{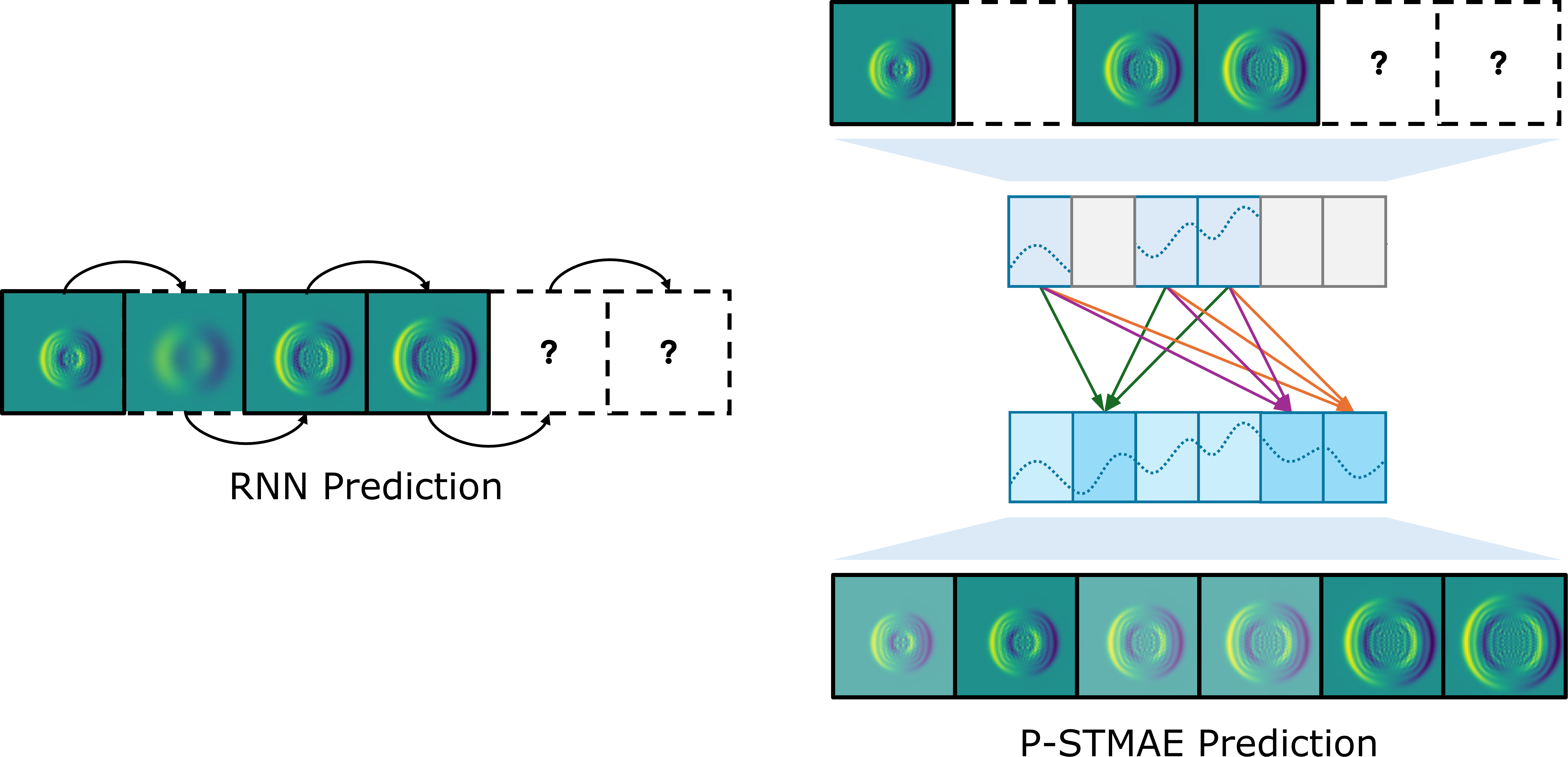}}
\caption{Comparison of Sequence-to-Sequence(Seq2Seq) prediction methods in dynamical systems of irregular time steps. \textbf{Left}: Traditional RNN-based models feature step-wise rolling out with necessary data imputation for handling missing steps, which may introduce biases and cumulative errors. \textbf{Right}: Our model performs element-wise predictions in the latent space by adaptive attention mechanism to reconstruct the complete sequence in a single pass.}
    \label{fig:introductory}
\end{figure*}

More recently, deep learning models like Convolutional Neural Networks (CNNs) and RNNs have shown advantages in surrogate modelling of time series problems by leveraging their ability to capture spatial and temporal patterns, respectively.
Deep Convolutional Recurrent Autoencoder (ConvRAE)~\cite{gonzalez2018deep} combines CNNs with Long Short-Term Memory Networks (LSTMs) to capture both spatial and temporal patterns.
However, it inherits RNNs' drawbacks, such as vanishing and exploding gradients~\cite{hochreiter1998vanishing}, which become detrimental for long sequences in  high-dimensional dynamical systems~\cite{chang2019antisymmetricrnn}.
Convolutional Long Short-Term Memory Network (ConvLSTM)~\cite{shi2015convolutional} combines convolutional operations with LSTM cells to directly model spatiotemporal relations.
Nevertheless, these RNN-based approaches rely on regularly sampled time series, limiting their applicability to irregular time steps.
Alternative methods~\cite{iakovlev2024modeling} require interpolation to estimate time points that do not align with the established time grid.
These limitations prompt the need for models capable of handling irregular time steps and incomplete data while preserving the integrity of physical processes (see Fig.~\ref{fig:introductory}).

Recent advancements in transformers~\cite{vaswani2017attention} have introduced a powerful architecture for time series modelling for handling irregular and incomplete data~\cite{geneva2022transformers,feichtenhofer2022masked}.
Originally developed for Natural Language Processing (NLP)~\cite{radford2019language}, transformers are well-suited for sequences of variable lengths and missing elements due to their self-attention mechanism.
Unlike RNNs that process inputs sequentially, this mechanism enables simultaneous attention to various parts of the input sequence and is highly parallelisable, thereby capturing long-range dependencies even when parts of the data are missing or unevenly spaced.
Methods like Bidirectional Encoder Representations from Transformers  (BERT)~\cite{devlin2018bert} in language understanding and Masked Autoencoder (MAE)~\cite{he2022masked} in image recognition demonstrate the efficacy of masking strategies for learning robust representations.
Building on these successes, Time Series Masked Autoencoder (TiMAE)~\cite{li2023ti} has demonstrated the utility of self-supervised learning and masked modelling for reconstructing missing data points in time series prediction.
However, this approach has largely been applied in low-dimensional systems, such as financial~\cite{karlsson2024detecting} or healthcare~\cite{patel2024emit} time series, and has not been fully extended to more complex, high-dimensional dynamical systems governed by physical processes.
Modelling these systems presents challenges due to computational complexity and memory usage associated with high dimensionality, while maintaining spatial and temporal coherence under irregular observations remains difficult for traditional deep learning models and common transformer variants~\cite{wen2022transformers}.

Existing approaches for irregular time series modelling in dynamical systems exhibit fundamental limitations.
Neural ordinary differential equation (Neural ODE)-based methods require continuous-time solvers and are often sensitive to stiffness and numerical instability in high-dimensional PDE systems.
Interpolation-based transformers and recurrent models rely on resampling or imputation, which can distort the true temporal dynamics.
In contrast, masked reconstruction enables direct modelling of irregularly sampled sequences without explicit interpolation or continuous-time integration.

To address these challenges, we propose a novel model called Physics Spatiotemporal Masked Autoencoder (P-STMAE).
Unlike previous transformer models primarily focused on low-dimensional data, P-STMAE is specifically designed for modelling high-dimensional dynamical systems, incorporating both spatial and temporal dependencies in a unified framework.
The core innovation lies in combining a convolutional autoencoder for spatial feature extraction with a masked autoencoder optimised for irregular time series prediction.
The convolutional autoencoder compresses high-dimensional physical data into a low-dimensional latent space, thereby reducing computational complexity while retaining essential spatiotemporal features.
In the latent space, a masked autoencoder uses the transformer's self-attention mechanism to predict future states.
The framework introduces placeholder and masking strategies to handle temporal dependencies among partially observed sequences, with positional encodings preserving the temporal order under irregular sequences.
The training adopts a purely data-driven approach~\cite{zhu2025nonlinear}, optimising a combination of physical and latent space losses without domain-specific knowledge.

We conduct numerical experiments on three datasets: two simulated scenarios from PDEBench~\cite{takamoto2022pdebench} (Shallow Water equations and Diffusion Reaction equations) and one real-world ocean fluid dataset, NOAA Sea Surface Temperature (SST)~\cite{huang2021improvements}, obtained from satellite and ship-based observations.
This combination ensures that P-STMAE adheres to scientific standards and generalises to diverse challenges in physical systems.

In summary, this paper makes the following key contributions:

\begin{itemize}
    \item A spatiotemporal masked autoencoder for latent dynamics modelling.
    \item Placeholder-based attention for handling irregular and missing time steps.
    \item A unified framework for sequence reconstruction and forecasting.
    \item The proposed model outperforms ConvLSTM and ConvRAE with improved efficiency and interpretability.
\end{itemize}

Our approach provides computational advantages over traditional physics-based PDE solvers.
Traditional time-stepping methods often require many small steps due to stability constraints and may involve iterative linear solves, whereas P-STMAE learns an approximation to the flow map and produces forecasts with a limited number of GPU-based forward passes~\cite{leveque2002fvmhp,cheng2025machine}, thereby reducing inference time and energy consumption while preserving spatiotemporal fidelity.

To avoid potential ambiguity, we clarify that the notion of "physical consistency'' in this work is relative rather than constraint-based.
P-STMAE does not explicitly enforce PDE residuals or conservation laws.
Instead, inspired by masked autoencoder approaches for time-series dynamics such as TS-MAE~\cite{liu2025ts}, it models the temporal evolution of physical fields in a latent space and improves the stability and coherence of the learned dynamics without introducing hard physics constraints.
Related latent-space forecasting approaches have also been explored and validated in independent studies~\cite{xu2025latent,riva2025comparison}.

To the best of our knowledge, our proposed model is among the first to unify Convolutional Autoencoder (CAE)-based spatial compression with masked temporal modelling using transformers in the latent space, specifically targeting high-dimensional, irregularly sampled dynamical systems.

The remainder of this paper is organised as follows: Section~\ref{ch:background} provides related work on reduced order modelling and deep learning-based approaches for high-dimensional dynamical systems.
Section~\ref{ch:methodology} introduces the proposed P-STMAE model, detailing its architecture and methodology.
Section~\ref{ch:experiments} presents the numerical experiments conducted on both synthetic and real-world datasets to evaluate the model's performance.
Finally, Section~\ref{ch:conclusion} discusses limitations and concludes the paper.

\section{Related work}\label{Xsec2-2}
\label{ch:background}

\subsection{Traditional approaches to irregular time series}\label{Xsec3-2.1}

Traditional approaches to handling temporally irregular observations include time series models like ARIMA~\cite{nelson1998time} and data assimilation algorithms such as Kalman Filters~\cite{gomez1994estimation}.
While ARIMA models are effective for univariate, stationary time series, they face limitations with high-dimensional, nonlinear systems~\cite{kontopoulou2023review,siami2018comparison}.
Kalman Filters estimate the state of a linear system from incomplete measurements but rely on assumptions of linearity and normality that often do not hold in complex, high-dimensional dynamics~\cite{carrassi2018data}.
Extended and Unscented Kalman Filters~\cite{ribeiro2004kalman,wan2000unscented} attempt to address nonlinearities but still struggle with high dimensionality and irregular time steps common in real-world data.
These limitations motivate the development of deep learning approaches capable of handling high-dimensional, nonlinear dynamics with irregular temporal sampling.

\subsection{Reduced order modelling and autoencoders}\label{Xsec4-2.2}

Reduced Order Modelling (ROM) aims to reduce the computational cost of simulating high-fidelity dynamical systems by constructing efficient surrogate models that preserve essential system dynamics~\cite{benner2015survey,asher2015review,fu2025parametric,pan2024domain,abbaszadeh2025reduced}.
These models enable faster predictions by approximating the original system in a lower-dimensional space while retaining relevant features for downstream tasks~\cite{fukami2023grasping,guo2025nonlinear}.

Traditional projection-based methods, such as Proper Orthogonal Decomposition (POD)~\cite{mackiewicz1993principal} or Dynamic Mode Decomposition (DMD), project data onto optimal linear subspaces that explain most of the variance.
However, their effectiveness is limited when the system exhibits strong nonlinearity and time-varying behaviour. Moreover, they are often intrusive, requiring access to the governing equations or system operators during the reduction process~\cite{karasozen2022intrusive}.

Deep learning offers a non-intrusive and flexible alternative for ROM.
In particular, autoencoders can learn nonlinear manifolds from data alone, making them effective in compressing and reconstructing high-dimensional spatial features~\cite{li2023comprehensive, wu2023deep}.
CAEs are widely used in spatiotemporal modelling, where spatial patterns can be compressed into latent representations and later decoded to reconstruct the original fields.

Formally, a CAE consists of an encoder and decoder:
\begin{align}
\mathbf{z}_t &= f_E(\mathbf{x}_t; \theta_E), \\
\hat{\mathbf{x}}_t &= f_D(\mathbf{z}_t; \theta_D),
\end{align}

where $t \in \mathbb{N}$ represents each valid time step, and $f_E: \mathbb{R}^{d_x} \rightarrow \mathbb{R}^{d_z}$ represents the encoding function that maps the input physical state $\mathbf{x}_t$ to the compressed latent space $\mathbf{z}_t$ using the parameters $\theta_E$, commonly $d_z \ll d_x$. Similarly, $f_D: \mathbb{R}^{d_z} \rightarrow \mathbb{R}^{d_x}$ denotes the decoding function that reconstructs the original physical state $\hat{\mathbf{x}}_t \in \mathbb{R}^{d_x}$ using the parameters $\theta_D$.

The training goal is to minimise the reconstruction loss, which measures the mean squared error between the input and output physical states averaged over total time steps $T$:
\begin{align}
\theta_E^*, \theta_D^* = \arg \min_{\theta_E, \theta_D} \frac{1}{T} \sum_{t=1}^T \mathbb{E} \left[ \left\|\hat{\mathbf{x}}_t - \mathbf{x}_t\right\|^2 \right].
\end{align}

A notable advantage of using deep autoencoders in ROM is their ability to learn nonlinear manifolds, for high-dimensional and nonlinear dynamics, which are common in fluid mechanics, climate systems, and biological simulations.

\subsection{Sequence modelling in latent space}\label{Xsec5-2.3}

RNNs~\cite{medsker2001recurrent} are a type of neural networks tailored for handling sequence data, making them effective for time series and sequential tasks. Unlike traditional feed-forward neural networks, RNNs capture information from previous states through internal hidden memories, enabling them to maintain and process temporal dependencies~\cite{lipton2015critical}. Specifically, LSTMs use three information gates, including input, forget, and output gates, to regulate what information should be added, retained, or output from the cell state, thus maintaining long-term dependencies more effectively~\cite{sundermeyer2012lstm}. They are widely used in reduced-order spatiotemporal system modelling in computational physics \cite{gong2025accelerating,cheng2023generalised}.

\subsubsection{Latent sequence forecasting via Conv{RAE} and Conv{LSTM}}\label{Xsec6-2.3.1}

One notable approach to employs RNNs for high-dimensional dynamical system modelling is the ConvRAE~\cite{gonzalez2018deep}. The model first employs a deep convolutional autoencoder to compress a sequence of physical fields $\{(\mathbf{x}_t, t)\}_{t=1}^T$ into the latent space $\{(\mathbf{z}_t, t)\}_{t=1}^T$. The compressed representation is then sent into an LSTM network to model its temporal evolution, thus preserving future predictive capabilities by autoregressive rollouts.

Mathematically, it predicts future states $\{(\hat{\mathbf{z}}_t, t)\}_{t=T_{\text{in}}+1}^{T_{\text{in}}+T_{\text{out}}}$ based on input states $\{(\mathbf{z}_t, t)\}_{t=1}^{T_{\text{in}}}$. For each forecasting step, it predicts the next state $\hat{\mathbf{z}}_{t+1}$ in an autoregressive manner based on the previous output $\hat{\mathbf{z}}_t$ and a hidden memory $\mathbf{h}_t$:
\begin{align}
\hat{\mathbf{z}}_{t+1}, \mathbf{h}_{t+1} = \text{LSTM-Cell}(\hat{\mathbf{z}}_t, \mathbf{h}_t).
\end{align}

Common loss such as Mean Squre Error (MSE) can be used to minimise the distance between predicted physical states and the ground truth, i.e.
\begin{align}
\hat{\mathbf{x}}_t &= f_D(\hat{\mathbf{z}}_t; \theta_D), \\
\mathcal{L}_{\text{ConvRAE}} &= \frac{1}{T_{\text{out}}} \sum_{t=T_{\text{in}}+1}^{T_{\text{in}}+T_{\text{out}}} \mathbb{E} \left[ \left\|\hat{\mathbf{x}}_t - \mathbf{x}_t\right\|^2 \right].
\end{align}

ConvRAE effectively integrates the strengths of CNNs in capturing localised spatial features and LSTMs in preserving temporal dependencies.
However, it inherits common drawbacks associated with RNNs, including susceptibility to vanishing and exploding gradients~\cite{hochreiter1998vanishing}, which become problematic when dealing with long sequences typical in high-dimensional dynamical systems~\cite{chang2019antisymmetricrnn}.

Another influential model is the ConvLSTM~\cite{shi2015convolutional}, which is specifically designed for spatiotemporal sequence forecasting in a Seq2Seq framework.
ConvLSTM extends the conventional Fully Connected LSTM (FC-LSTM) by incorporating convolutional operations directly into the cell of the LSTM to model temporal transitions at the full physical space.
Unlike ConvRAE, which operates as a two-stage method, ConvLSTM offers an end-to-end approach that simultaneously captures both spatial and temporal dependencies within a unified architecture.
However, both ConvRAE and ConvLSTM require regularly sampled time series, which limits their applicability to irregular time steps without preprocessing.

\subsubsection{Challenges with irregular time series}\label{Xsec7-2.3.2}

While RNNs and their variants excel in modelling regularly sampled time series, they face significant challenges when dealing with irregular time steps~\cite{lechner2020learning} and high dimensionality~\cite{yang2017breaking}. Standard RNN-based Seq2Seq models inherently assume consistent sampling intervals, making it difficult to handle missing or unevenly spaced observations without additional preprocessing steps.

To address this, many RNN-based models rely on explicit imputation or interpolation as a preprocessing step~\cite{weerakody2021review}, including statistical interpolation~\cite{fan2017spatiotemporal,johnson2022empirical}, nearest neighbour search~\cite{broersen2006estimating}, and time-aware data filling~\cite{lepot2017interpolation}.
However, these preprocessing procedures can introduce bias, increase computational cost, and obscure the true temporal dynamics of the system~\cite{afrifa2020missing,ahn2022comparison}, thereby limiting their effectiveness in capturing complex spatiotemporal dynamics under high sparsity.

More recent work has used generative models such as Variational Autoencoders (VAEs)~\cite{fortuin2020gp} and Generative Adversarial Networks (GANs)~\cite{guo2019data,luo2018multivariate} to impute missing values. However, these approaches may suffer from data inefficiency, instability in long-term predictions, and non-uniqueness of outputs, limiting their use in physical system modelling~\cite{fang2020time, yoon2018gain, fortuin2020gp}.
While probabilistic generative models are highly effective for noisy or underdetermined real-world systems and can explicitly represent predictive uncertainty~\cite{zhuang2025spatially}, this work focuses on deterministic reconstruction to enable clarity and controlled analysis of irregularly sampled latent dynamics in deterministic PDE simulation settings, where the governing equations define a single-valued flow map~\cite{riva2025data}.

\subsection{Masked transformers and physics-informed latent modelling}\label{Xsec8-2.4}

Existing approaches for irregular time series modelling in dynamical systems exhibit fundamental limitations.
Neural ordinary differential equation (Neural ODE) based methods require continuous-time solvers and are often sensitive to stiffness and numerical instability in high-dimensional PDE systems.
Interpolation-based transformers and recurrent models rely on resampling or imputation, which can distort the true temporal dynamics.
Masked reconstruction offers an alternative by enabling direct modelling of irregularly sampled sequences without explicit interpolation or continuous-time integration.

Recent transformer-based models have shown promise in modelling long-range dependencies and handling irregular sequences directly through masked attention and learned time encodings.
Works like Time-MAE and masked sequence models~\cite{gao2022convmae, zerveas2021transformer} have successfully addressed missing data without interpolation.
Building on these successes, TiMAE~\cite{li2023ti} has demonstrated the utility of self-supervised learning and masked modelling for reconstructing missing data points in time series prediction.
However, these approaches have primarily been applied to low-dimensional systems, such as financial~\cite{karlsson2024detecting} or healthcare~\cite{patel2024emit} time series, and have not been fully extended to the more complex, high-dimensional dynamical systems governed by physical processes.
Modelling these systems presents challenges due to computational complexity and memory usage associated with high dimensionality, while maintaining spatial and temporal coherence under irregular observations remains difficult for traditional deep learning models and common transformer variants~\cite{wen2022transformers}.

Simultaneously, there is a growing interest in combining physical priors with latent representation learning~\cite{fu2023physics}. Physics-informed neural networks (PINNs) and hybrid data-physics models have shown effectiveness in preserving physical consistency and improving generalisation under small data regimes.

\section{Methodology}\label{Xsec9-3}
\label{ch:methodology}

P-STMAE addresses irregular time series prediction using a masked modelling strategy by combining a convolutional autoencoder for spatial representation and a masked transformer for temporal modelling.
The framework directly handles irregular time steps through masked reconstruction, eliminating the need for preprocessing or regular sampling that can introduce bias and computational overhead.
By leveraging transformer self-attention instead of sequential RNNs processing, P-STMAE captures long-range dependencies without suffering from vanishing gradients or requiring regular input sequences.
In contrast to Neural ODE methods that require continuous-time solvers, P-STMAE learns a discrete flow map in a latent space, enabling efficient inference without numerical integration.
To address the computational challenges of applying transformers to high-dimensional systems, P-STMAE operates in a compressed latent space and reduces memory usage while maintaining spatiotemporal coherence.
This section outlines the overall framework, the encoder and decoder designs, the attention-based temporal model, and the training loss formulation.

\subsection{Overall framework}\label{Xsec10-3.1}

Mathematically, we consider a physical sequence consisting of input states $\mathbf{X}_{T_{\text{in}}}$ defined on input steps $T_{\text{in}} = \{1, 2, \ldots , t_{\text{in}}\}$, followed by forecasting states $\mathbf{X}_{T_{\text{out}}}$ on output steps $T_{\text{out}} = \{t_{\text{in}}+1, \ldots , t_{\text{in}}+t_{\text{out}}\}$. We denote:
\begin{align}
\mathbf{X}_{T_{\text{in}}} = \{(\mathbf{x}_t, t) \mid t \in T_{\text{in}}\}, \quad
\mathbf{X}_{T_{\text{out}}} = \{(\mathbf{x}_t, t) \mid t \in T_{\text{out}}\},
\end{align}
where $\mathbf{x}_t \in \mathbb{R}^{d_x}$ is the physical state at time $t$.

To simulate irregularity, we randomly split $T_{\text{in}}$ into two disjoint sets: $T_{\text{obs}}$ (observed steps) and $T_{\text{miss}}$ (missing steps), such that $T_{\text{obs}} \cup T_{\text{miss}} = T_{\text{in}}$.
This design enables learning from partially observed sequences without data imputation or interpolation, thereby addressing the limitations inherent in preprocessing-based approaches.
Missing and future steps are replaced with placeholder variables $\Phi_x$:
\begin{align}
\Phi_x = \{(\phi_x, t) \mid t \in T_{\text{miss}} \cup T_{\text{out}}\}, \quad \phi_x \in \mathbb{R}^{d_x}.
\label{eq:phy_placeholders}
\end{align}

The model then reconstructs the complete sequence $\hat{\mathbf{X}}$ from $\mathbf{X}_{T_{\text{obs}}}$ and placeholders $\Phi_x$:
\begin{align}
\hat{\mathbf{X}} = \text{P-STMAE}(\mathbf{X}_{T_{\text{obs}}}, \Phi_x) = \{(\hat{\mathbf{x}}_t, t) \mid t \in T_{\text{in}} \cup T_{\text{out}}\}.
\end{align}

\begin{figure*}[!t]

\centering
    \subfloat[Encoder pipeline]{
        \includegraphics[width=0.32\textwidth]{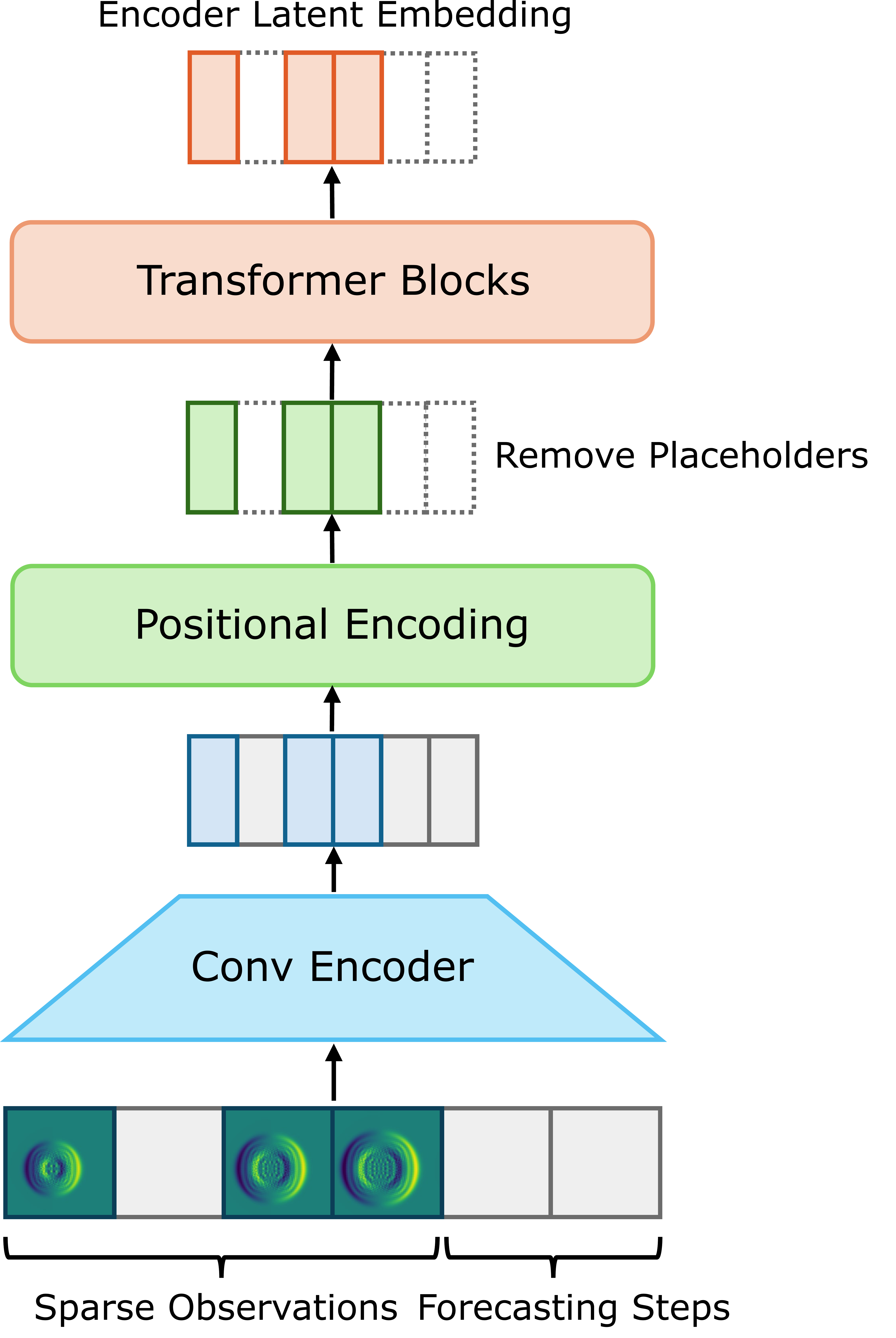}
        \label{fig:model_architecture_left}
    }
    \subfloat[Decoder pipeline]{
        \includegraphics[width=0.32\textwidth]{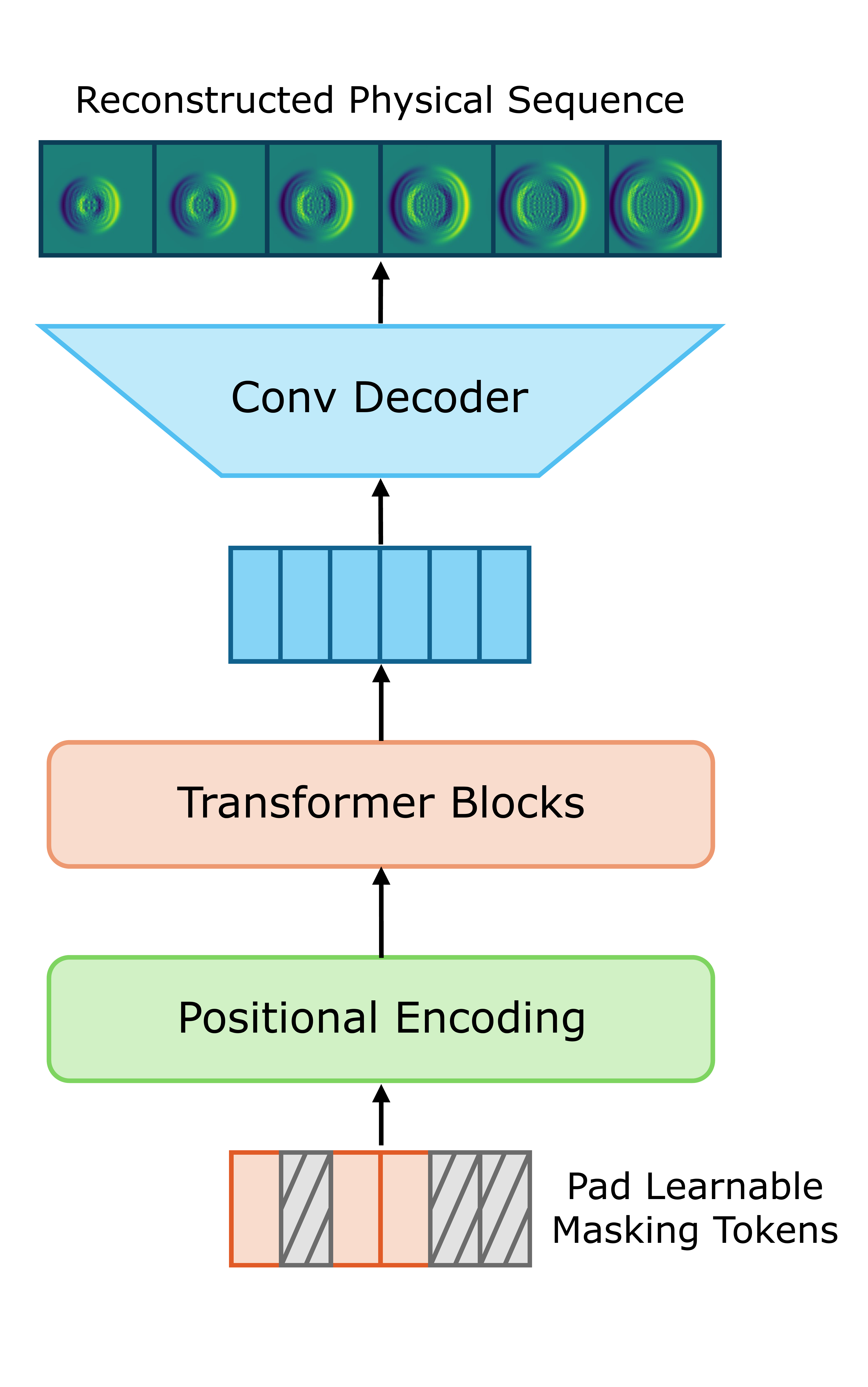}
        \label{fig:model_architecture_right}
    }
    \subfloat[Transformer block]{
        \includegraphics[width=0.32\textwidth]{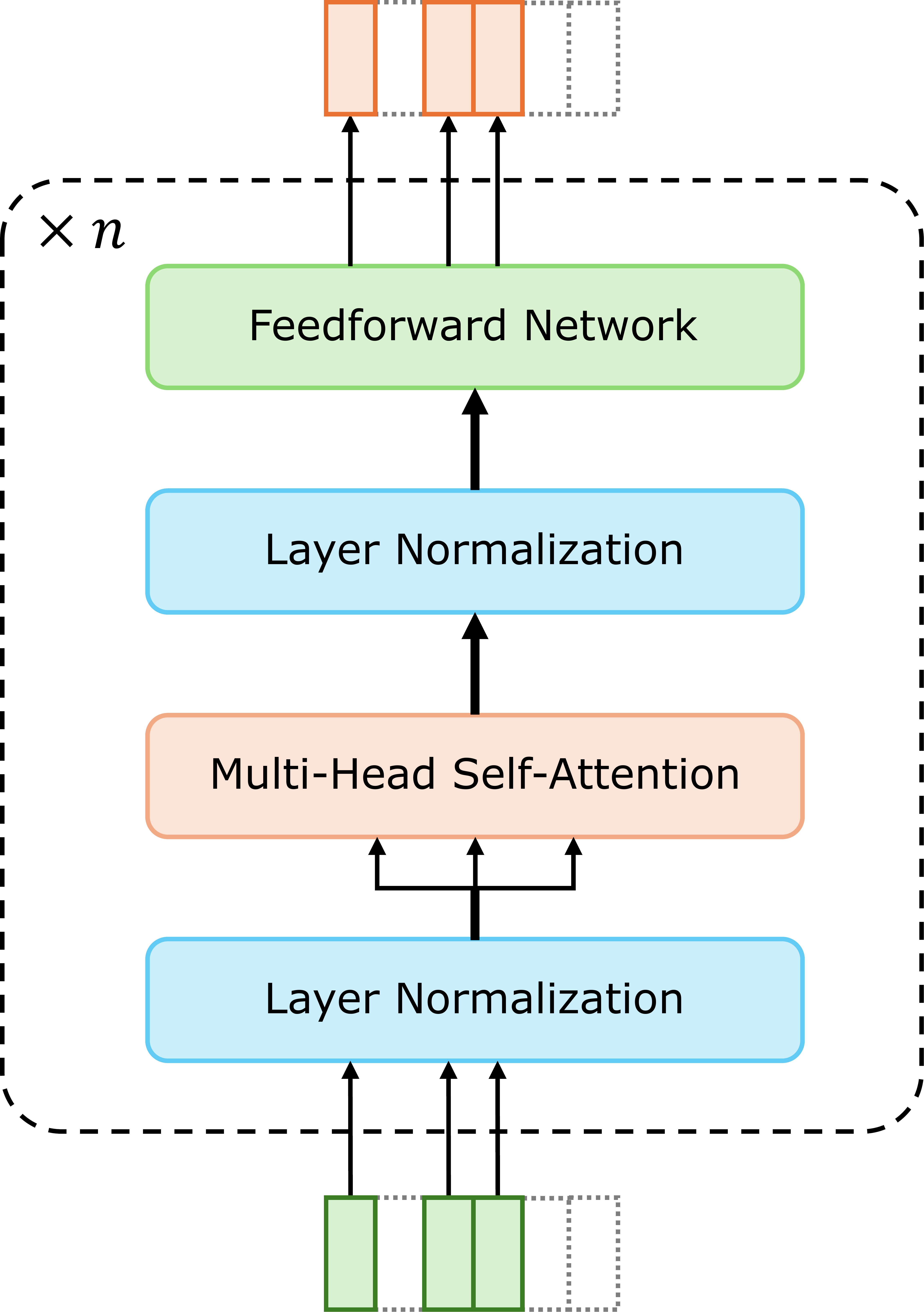}
        \label{fig:transformer_block}
    }
\caption{
        Architecture of the proposed P-STMAE framework.
        \textbf{(a)} Convolutional encoder compresses physical states into latent representations.
        Positional encodings are added, and a masked transformer captures temporal dependencies in latent space.
        \textbf{(b)} Learnable masking tokens are padded at missing and future time steps.
        Transformer blocks process the sequence, and the convolutional decoder reconstructs the complete physical fields.
        \textbf{(c)} Each transformer block consists of layer normalization, multi-head self-attention, and a feedforward network.
        Self-attention operates only on observed latent states.
    }
    \label{fig:model_architecture}
\end{figure*}

Fig.~\ref{fig:model_architecture} illustrates the overall pipeline of the proposed P-STMAE framework.

\subsection{Spatial encoder: Convolutional autoencoder}\label{Xsec11-3.2}

To reduce spatial redundancy and improve computational efficiency, we adopt a CAE to project high-dimensional inputs $\mathbf{x}_t \in \mathbb{R}^{d_x}$ into compact latent representations $\mathbf{z}_t \in \mathbb{R}^{d_z}$, where $d_z \ll d_x$.
This compression mitigates the computational complexity and memory challenges of applying transformers directly to high-dimensional physical fields, facilitating efficient temporal modelling in the latent space.

During encoding, physical placeholders $\Phi_x$ are converted into latent placeholders $\Phi_z$:
\begin{align}
\Phi_z = \{(\phi_z, t) \mid t \in T_{\text{miss}} \cup T_{\text{out}}\}, \quad \phi_z \in \mathbb{R}^{d_z}.
\end{align}
These placeholders are excluded from backpropagation and later masked out in the transformer blocks.

The placeholder states $\phi_x$ and $\phi_z$ are fixed tensors (set to zero after normalisation), excluded from gradient backpropagation, and serve solely as positional anchors for masking and attention mechanisms.

\subsection{Temporal modelling: Masked autoencoder }\label{Xsec12-3.3}

The masked autoencoder learns latent temporal dynamics from partially observed sequences.
By operating directly on irregularly sampled latent states without preprocessing, this approach circumvents the bias and distortion that interpolation methods typically introduce.
It processes the latent inputs $\mathbf{Z}_{T_{\text{obs}}} = \{(\mathbf{z}_t, t) \mid t \in T_{\text{obs}}\}$ using transformer blocks.

\subsubsection{Masked transformer blocks}\label{Xsec13-3.3.1}

Transformer blocks (see Fig.~\ref{fig:model_architecture}(c)) compute self-attention over observed latent states.
In contrast to RNNs that process sequences sequentially and are prone to vanishing gradients, self-attention enables parallel processing and direct access to all observed time steps, capturing long-range dependencies without gradient degradation.
Given the hidden and attention dimension, the attention weights are computed as:
\begin{align}
\mathbf{Q} = \mathbf{Z}_{T_{\text{obs}}} \mathbf{W}_Q, \quad 
\mathbf{K} = \mathbf{Z}_{T_{\text{obs}}} \mathbf{W}_K, \quad 
\mathbf{V} = \mathbf{Z}_{T_{\text{obs}}} \mathbf{W}_V,
\end{align}
\begin{align}
\mathbf{A} = \text{softmax}\left(\frac{\mathbf{Q} \cdot \mathbf{K}^T}{\sqrt{d_k}}\right), \quad 
\mathbf{O} = \mathbf{A} \cdot \mathbf{V}, \quad 
\mathbf{O'} = \text{Linear}(\mathbf{O}) + \mathbf{Q}.
\end{align}

\subsubsection{Decoder and masking tokens}\label{Xsec14-3.3.2}

The decoder uses a lighter transformer stack to reconstruct the full latent sequence $\hat{\mathbf{Z}} = \{(\hat{\mathbf{z}}_t, t) \mid t \in T_{\text{in}} \cup T_{\text{out}}\}$ by attending to encoded representations and learnable masks on missing steps.
This enables parallel, non-autoregressive prediction, which eliminates the error accumulation inherent in autoregressive RNNs and facilitates efficient single-pass inference.

\subsubsection{Positional embeddings}\label{Xsec15-3.3.3}

We inject sine-cosine positional embeddings $\delta_t$ into all latent inputs:
\begin{align}
\delta(t, 2i) = \sin\left(\frac{t}{10000^{\frac{2i}{d_z}}}\right), \quad \delta(t, 2i+1) = \cos\left(\frac{t}{10000^{\frac{2i}{d_z}}}\right), \\
\mathbf{z}_t \leftarrow \mathbf{z}_t + \delta_t, \quad t \in T_{\text{in}} \cup T_{\text{out}}
\end{align}
where $\delta_t$ is non-trainable and ensure temporal consistency across irregular steps.

We note that the sinusoidal positional encoding is used solely to encode relative temporal ordering among irregular time steps, rather than representing absolute physical time scales.

Unlike RNN-based models, our masked transformer supports single-step inference over the entire sequence, reducing latency and eliminating autoregressive error accumulation.

This masked reconstruction paradigm is advantageous for irregular time series.
Rather than relying on explicit interpolation or resampling that may distort temporal dynamics, the model learns to infer missing or unevenly spaced observations directly from surrounding context.
By training on partially observed sequences, P-STMAE naturally develops robustness to irregular sampling patterns.
Operating in the latent space further enhances this capability, as the model can exploit global spatiotemporal dependencies while avoiding artifacts introduced by preprocessing, thereby addressing the computational challenges of high-dimensional systems.
This explains why masked autoencoders demonstrate stronger adaptability in irregular settings compared to RNN-based approaches that require regularised inputs.

\subsection{Loss function and training objective}\label{Xsec16-3.4}

We jointly minimise the prediction errors in both physical and latent spaces:
\begin{align}\label{eq:loss}
\mathcal{L}
= \frac{1}{T_{\text{in}} + T_{\text{out}}} \sum_{t=1}^{T_{\text{in}} + T_{\text{out}}} \left( \mathbb{E}\left[ \left\|\hat{\mathbf{x}}_t - \mathbf{x}_t\right\|^2 \right] + \lambda \cdot \mathbb{E}\left[ \left\|\hat{\mathbf{z}}_t - \mathbf{z}_t\right\|^2 \right] \right),
\end{align}
where $\lambda$ balances physical and latent consistency.
Both terms in Eq.~(\ref{eq:loss}) employ the L2 norm (mean squared error), which is a standard choice for continuous physical fields and provides smooth, stable gradients for training masked autoencoder architectures.

\subsection{Evaluation metrics}\label{Xsec17-3.5}

To comprehensively evaluate model performance, we employ three complementary metrics commonly used in spatiotemporal forecasting~\cite{takamoto2022pdebench,gonzalez2018deep}:
The MSE measures pixel-wise prediction accuracy over all time steps~\cite{mahalakshmi2016survey}.
The Structural Similarity Index Measure (SSIM) quantifies structural similarity of spatial fields, capturing perceptual quality and spatial coherence~\cite{wang2004image}.
The Peak Signal-to-Noise Ratio (PSNR) assesses reconstruction fidelity in decibels, providing a measure of signal-to-noise ratio~\cite{huynh2012accuracy}.
These metrics collectively capture pointwise accuracy, structural preservation, and reconstruction quality, which are essential for evaluating spatiotemporal predictions in physical systems.

\subsection{Ablation study design}\label{Xsec18-3.6}

We compare P-STMAE against two established RNN-based models to evaluate the effectiveness of transformer-based latent modelling versus recurrent approaches:
\begin{enumerate}
\item ConvRAE~\cite{gonzalez2018deep} employs a two-stage approach: a convolutional autoencoder for spatial compression followed by LSTM for temporal modelling in the latent space.
This enables a direct comparison of transformer-based versus LSTM-based temporal modelling within the same latent representation framework.
\item ConvLSTM~\cite{shi2015convolutional} integrates convolutional operations with LSTM cells to model spatiotemporal relations directly in full physical space.
This provides a comparison with end-to-end full-space approaches, contrasting with the latent-space methods used by ConvRAE and P-STMAE.
\end{enumerate}

To ensure fair comparison, we train these RNN-based baselines using ground truth inputs during training and autoregressive predictions during inference.
Since RNN-based models require regularly sampled inputs, we adapt them to handle irregular time steps using linear interpolation, following common practice in irregular time series modelling~\cite{yoon2018estimating,che2018recurrent}.
This preprocessing step allows the baselines to process the data while enabling comparison with P-STMAE, which handles irregular sampling directly without interpolation.

\section{Experiments}\label{Xsec19-4}
\label{ch:experiments}

\subsection{Overview}\label{Xsec20-4.1}

\subsubsection{Datasets and benchmarking}\label{Xsec21-4.1.1}

Following the methodology outlined in Section~\ref{ch:methodology}, we evaluate P-STMAE against the baseline models introduced in the Ablation Study Design on three representative datasets.
These datasets span synthetic PDE simulations and real-world climate observations, enabling validation of both accuracy under controlled conditions and generalisation to noisy, large-scale data.
Specifically, we use two datasets from PDE simulations and one from real-world observations:
\begin{itemize}
    \item Shallow Water~\cite{cheng2019background}: nonlinear fluid flow, testing robustness to chaotic dynamics.
    \item Diffusion Reaction from PDEBench~\cite{takamoto2022pdebench}: chemical patterns, testing coupled-variable modelling.
    \item SST~\cite{huang2021improvements}: NOAA sea surface temperature data, which are noisy data with long-range dependencies.
\end{itemize}

Each dataset is split into training, validation, and test sets with ratios of $(0.8, 0.1, 0.1)$.
Channel-wise normalisation is applied to transform data values into the range of $[0, 1]$ prior to both training and evaluation. This ensures a consistent dynamic range across heterogeneous physical variables.

We use a shifting window approach to sample input sequences from the original dataset, which contains longer sequences generated from simulations.
This approach allows the model to fully leverage all available input sequences, enhancing its adaptability to various prediction scenarios.
The input length is $T_{\text{in}} = 10$, and the forecasting length is $T_{\text{out}} = 5$ for all datasets.
We emphasise that this choice of forecasting length is made solely for fair comparison with ConvRAE and ConvLSTM baselines and does not reflect a limitation of the proposed model. In P-STMAE, future time steps are treated as masked positions in the latent space and reconstructed in a non-autoregressive manner. As a result, the forecasting horizon can be flexibly adjusted by specifying a different set of masked future time indices, without any change to the model architecture or training procedure.
To model the irregular time series, missing steps are sampled with a random mask for each input sequence with a missing ratio of 0.5, except for Section~\ref{sec:sw_missing_ratio} which uses mixed ratios, consistent between training and evaluation.

We compare P-STMAE with two representative models: ConvRAE, which also relies on latent representations with RNN temporal modelling, and ConvLSTM, which performs full-space sequence learning.
This ensures a fair comparison between latent-space transformer, latent-space RNN, and full-space RNN-based approaches.
Evaluation metrics include:
\begin{itemize}
\item MSE: pixel-wise accuracy,
\item SSIM: preservation of structural information,
\item PSNR: reconstruction fidelity and noise robustness.
\end{itemize}

All metrics are computed on the normalised fields. In particular, SSIM and PSNR are evaluated using their standard definitions after channel-wise normalisation, with the dynamic range parameter for PSNR set to $\mathrm{MAX}=1$. This avoids unit dependence on the original physical variables and enables consistent comparison across datasets~\cite{takamoto2022pdebench,gonzalez2018deep,geneva2022transformers}.

\subsubsection{Implementation details}\label{Xsec22-4.1.2}

Since both P-STMAE and ConvRAE use a CAE combined with a latent model structure, our goal is to compare their performance in latent space inference using the same spatial autoencoder.
To achieve this, we pre-trained an optimal CAE on the training dataset, then froze its parameters and used it with both models in subsequent time series experiments.
The latent dimension is set to 128 for all datasets.
To maintain consistency between predictions and ground truth during evaluation, a Sigmoid activation function is applied at the output of the CAE decoder, ensuring that all reconstructed fields lie strictly within the normalised range of $[0, 1]$.

Both the encoder and decoder transformer blocks possess 2 heads, with the encoder having a depth of 4 and the decoder a depth of 1.
The positional embedding settings follow those of the original Transformer architecture~\cite{vaswani2017attention}.
An exception is made for SST data, where the model is expanded to include 8 heads and an encoder depth of 8.
We use the RAdam optimiser with a learning rate of $3 \times 10^{-4}$ for training P-STMAE, and use the Adam optimiser with a learning rate of $1 \times 10^{-3}$ for training RNN baselines.
The batch size is 32.
The weighting coefficient of the combined loss is set to $\lambda = 0.5$ as shown in Eq.~(\ref{eq:loss}).

A common concern is that Transformer-based models require extremely large datasets, an observation that primarily arises from applications in natural language processing and natural-image modelling, where data distributions are highly complex and high-entropy~\cite{kaplan2020scaling,radford2019language,dosovitskiy2020image}.

In contrast, the scientific computing problems studied here are governed by smooth, structured PDE dynamics that lie on low-dimensional manifolds. Together with latent-space compression via the convolutional autoencoder, this reduces temporal modelling complexity. As a result, dataset sizes of $10^5$--$10^6$ frames are sufficient in our setting, and the Transformer-based temporal model does not exhibit data inefficiency compared to ConvLSTM.

The architecture is shown in Appendix Table~\ref{tab:cae_structure}.
All convolutions use kernel size $3 \times 3$ with same padding.

Table~\ref{tab:combined_test_metrics} presents a comparison of test performance metrics across all three datasets.
The results demonstrate that P-STMAE consistently achieves competitive or better performance compared to baseline models.
On the Shallow Water dataset, P-STMAE outperforms both ConvRAE and ConvLSTM across all metrics, achieving the lowest MSE ($6.16 \times 10^{-5}$), highest SSIM (0.9538), and highest PSNR (43.90).
For the Diffusion-Reaction dataset, P-STMAE achieves the lowest MSE ($5.99 \times 10^{-5}$) while ConvLSTM performs slightly better in SSIM and PSNR.
On the real-world SST dataset, P-STMAE delivers the strongest overall performance, outperforming both baselines with the lowest MSE ($8.02 \times 10^{-5}$), highest SSIM (0.9817), and highest PSNR (41.03).
These results indicate the robustness and generalisation capability of P-STMAE across diverse spatiotemporal systems, from synthetic PDE simulations to real-world climate observations.

\begin{table}
        \caption{Test metrics in the full physics space across all three datasets with a missing ratio of 0.5. Metrics for Shallow Water are averaged over all $h$, $u$, and $v$ variables; metrics for Diffusion-Reaction are averaged over both $u$ and $v$ variables. Bold values indicate the best performance among the three forecasting models (excluding CAE).}
    \begin{tabular}{lllll}
\toprule
        {Dataset} & {Model} & {MSE}               & {SSIM} & {PSNR} \\
        \midrule
        \multirow{4}{*}{Shallow Water}
                            & \textit{CAE}   & $5.32 \times 10^{-5}$      & $0.9596$      & $44.64$       \\
                            & P-STMAE        & $\bm{6.16 \times 10^{-5}}$ & $\bm{0.9538}$ & $\bm{43.90}$  \\
                            & ConvRAE        & $9.86 \times 10^{-5}$      & $0.9394$      & $42.47$       \\
                            & ConvLSTM       & $1.82 \times 10^{-4}$      & $0.9231$      & $40.72$       \\
        \multirow{4}{*}{Diffusion-Reaction}
                            & \textit{CAE}   & $3.66 \times 10^{-5}$      & $0.9887$      & $48.78$       \\
                            & P-STMAE        & $\bm{5.99 \times 10^{-5}}$ & $0.9870$      & $44.20$       \\
                            & ConvRAE        & $8.48 \times 10^{-5}$      & $0.9875$      & $44.41$       \\
                            & ConvLSTM       & $6.80 \times 10^{-5}$      & $\bm{0.9928}$ & $\bm{47.36}$  \\
        \multirow{4}{*}{SST}
                            & \textit{CAE}   & $7.11 \times 10^{-5}$      & $0.9819$      & $41.53$       \\
                            & P-STMAE        & $\bm{8.02 \times 10^{-5}}$ & $\bm{0.9817}$ & $\bm{41.03}$  \\
                            & ConvRAE        & $1.03 \times 10^{-4}$      & $0.9803$      & $40.33$       \\
                            & ConvLSTM       & $4.57 \times 10^{-4}$      & $0.9384$      & $36.84$       \\
    \bottomrule
\end{tabular}
    \label{tab:combined_test_metrics}
\end{table}

\subsection{Shallow water test case}\label{Xsec23-4.2}

\subsubsection{Dataset description}\label{Xsec24-4.2.1}

Shallow Water Equations (SWEs) are a set of hyperbolic PDEs that model the flow beneath a pressure surface in a fluid.
This dataset tests the models' ability to capture chaotic nonlinear fluid dynamics under varying physical parameters, which are relevant to geophysical flows such as atmospheric and oceanic dynamics.
The system is formulated as follows:
\begin{align}
    & \frac{\partial h}{\partial t} + \frac{\partial (hu)}{\partial x} + \frac{\partial (hv)}{\partial y} = 0, \\
    & \frac{\partial u}{\partial t} + g \frac{\partial h}{\partial x} + b u = 0, \\
    & \frac{\partial v}{\partial t} + g \frac{\partial h}{\partial y} + b v = 0,
\end{align}

where $h$ is the surface height of water, $u$ and $v$ are the orthogonal velocity components averaged in depth, $g$ is the gravitational acceleration, and $b$ is the friction coefficient of the fluid.
The initial condition is a cylinder bump in the water with a small height $h$ above the surface average, with a variable radius $r$, and zero velocities $u$ and $v$.
The boundary conditions are periodic in both directions.

Our numerical simulation is performed on a grid of size $128 \times 128$ with three variables $[h, u, v]$.
We sample snapshots from the simulation at fixed intervals when generating each data sequence.
The fixed time step is set to $\Delta t = 10^{-4}$, space step $\Delta x = 10^{-2}$, and gravity $g = 1.0$.
Other parameters, including the fluid friction $b$ and the centre, radius, and height of the cylinder bump, and the snapshot gap, are randomised to produce different initial conditions of fluid dynamics.
The detailed ranges of simulation parameters are shown in Appendix Table~\ref{tab:sw_parameters}.
We generate 600 sequences, with each sequence having 200 spatio-temporal frames.

\subsubsection{CAE reconstruction}\label{Xsec25-4.2.2}

First, we train a convolutional autoencoder on the shallow water dataset.
The CAE can capture the spatial features of input physical fields and reconstruct them with high fidelity.
It achieves an MSE of $5.32 \times 10^{-5}$, SSIM of 0.9596, and PSNR of 44.64 on the test set (see Appendix Fig.~\ref{fig:sw_cae_evaluation}).
These strong reconstruction metrics indicate that the CAE provides reliable latent representations for subsequent sequence modelling.

\subsubsection{Model comparison}\label{Xsec26-4.2.3}

The validation results for the shallow water dataset demonstrate the efficacy of the P-STMAE model compared to the baseline models.
In terms of full-space MSE, the P-STMAE outperforms both ConvRAE and ConvLSTM.
The full-space MSE curve for P-STMAE lies much closer to that of the CAE's performance, indicated by the dotted line (see Appendix Fig.~\ref{fig:sw_comparison}). 
This proximity suggests that the time series masked autoencoder is effective at exploiting the semantic information hidden in latent representations, making P-STMAE's performance nearly match the reconstruction capability of the CAE.
Analysing the latent MSE curves, we observe a consistent pattern whereby the P-STMAE surpasses the ConvRAE in terms of reducing prediction error within the latent space.
The performance indicates an ability to handle complex nonlinear spatiotemporal patterns, which the baseline models struggle with.
This confirms that transformer-based latent inference is more effective than RNN-based alternatives in the presence of chaotic fluid dynamics.

Despite its performance, the P-STMAE exhibits a relatively slower convergence speed compared to the RNN-based models.
This behaviour can be attributed to the complex self-attention mechanisms within the transformer architecture, which require a slower learning process to adequately tune weights to capture overall time series dependencies.

\begin{figure*}[!t]

\centerline{\includegraphics[width=\textwidth]{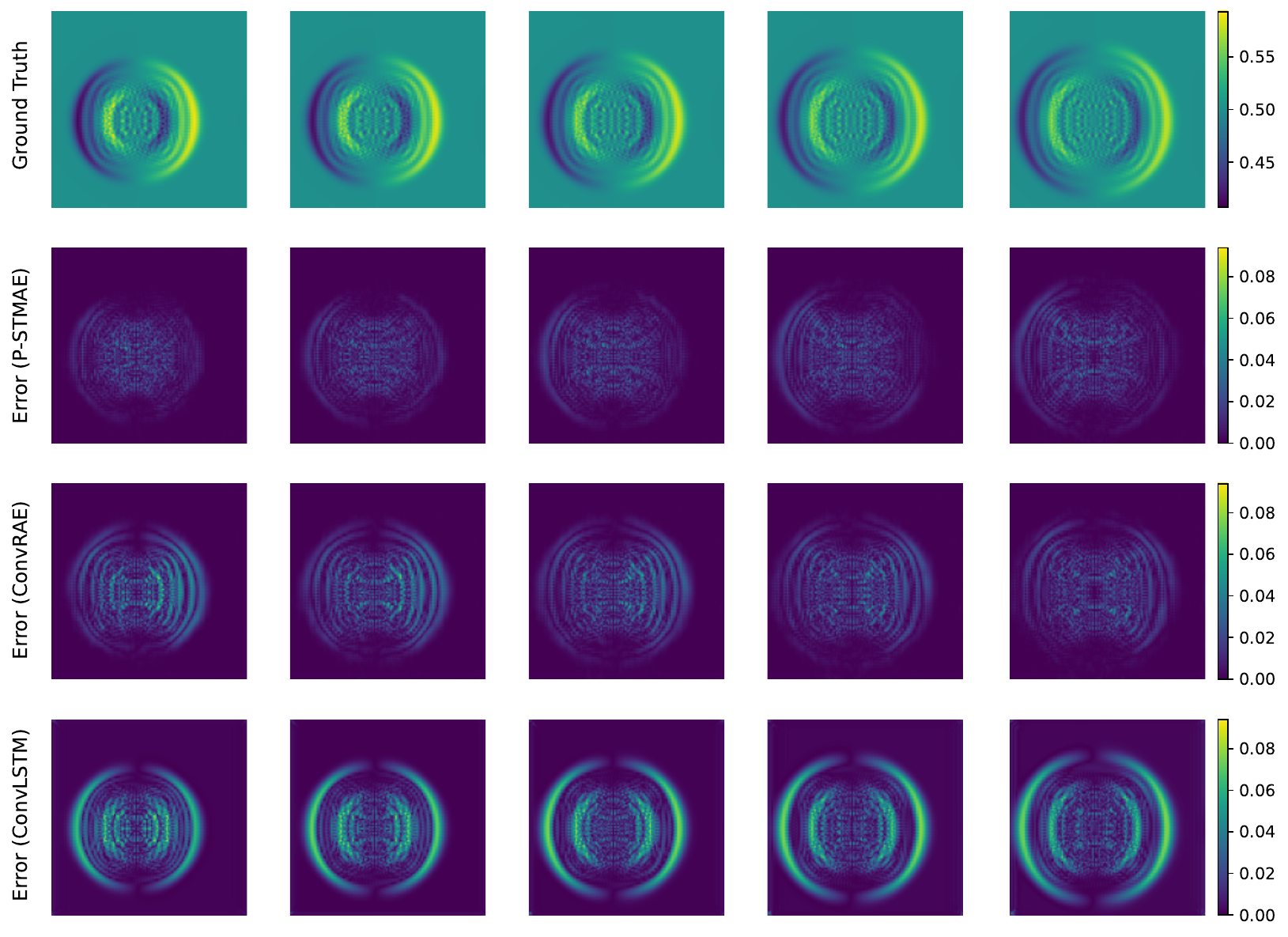}}
\caption{Ground truth (top) and error maps of P-STMAE, ConvRAE, and ConvLSTM for forecasting the variable $u$ in the shallow water dataset with a sampling dilation of 3. Columns represent successive forecasting steps. Among the models, P-STMAE yields the smallest errors, indicating predictive accuracy.}
    \label{fig:sw_err_comparison}
\end{figure*}

The test performance metrics for the shallow water dataset (see Table~\ref{tab:combined_test_metrics}) further substantiate the superior performance of the P-STMAE model over the baseline models.
Apart from the lowest full-space MSE, the P-STMAE demonstrates the highest SSIM, reflecting its ability to maintain structural integrity and perceptual quality of the reconstructed physical fields.
In terms of PSNR, the P-STMAE also scores the highest, confirming its effectiveness in minimising noise and enhancing the clarity of predictions compared to the other models.
Fig.~\ref{fig:sw_err_comparison} provides a detailed visual comparison of prediction errors across all three models.
The error maps reveal that P-STMAE consistently produces the smallest prediction errors across successive forecasting steps, with error magnitudes lower than both ConvRAE and ConvLSTM.
The spatial distribution of errors for P-STMAE is more uniform and concentrated in regions with higher physical complexity, while the baseline models exhibit larger and more widespread error patterns in areas with strong nonlinear dynamics (see Appendix Fig.~\ref{fig:sw_pred_err_comparison}).

\subsubsection{Ablation study on loss weighting coefficient}\label{Xsec27-4.2.4}

We conduct an ablation study on the shallow water dataset to investigate the sensitivity of P-STMAE to the weighting coefficient $\lambda$ in the combined loss function (Eq.~\ref{eq:loss}).
The ablation study results demonstrate that model performance is not sensitive to $\lambda$ within the range $[0.2, 1.0]$ (see Appendix Table~\ref{tab:lambda_ablation} and Fig.~\ref{fig:lambda_ablation}).
Based on these findings, we fix $\lambda = 0.5$ for all experiments for simplicity, as this value provides a balanced trade-off between physical and latent space consistency while maintaining robust performance across different $\lambda$ values.

\subsubsection{Missing ratio analysis}\label{Xsec28-4.2.5}
\label{sec:sw_missing_ratio}

We analyse model performance under varying amounts of missing data to assess generalisation capability as the number of missing steps increases.
To evaluate their performance, we train all models with missing steps ranging from 1 to 6 with a fixed input sequence length of 10 and evaluate them under corresponding settings.
During training, we maintain a fixed missing ratio within each batch and randomise this ratio between different batches to leverage parallelism and accelerate training.

Fig.~\ref{fig:sw_robustness} presents the results of these robustness experiments.
Panel (a) shows performance comparison under varying numbers of missing steps, while panel (b) shows test performance comparison regarding sampling dilations.
For missing ratio analysis (Fig.~\ref{fig:sw_robustness}(a)), P-STMAE consistently outperforms the baseline models in MSE across all missing step conditions.
The error curves demonstrate that P-STMAE maintains consistently low prediction errors even as the number of missing steps increases from 1 to 6, with the MSE curve remaining relatively flat.
In contrast, both ConvRAE and ConvLSTM show progressively increasing errors as missing steps increase, with ConvLSTM exhibiting sharp degradation.
This indicates that the transformer-based architecture of P-STMAE, which leverages attention mechanisms and contextual encoding, maintains lower prediction errors when reconstructing sequences with varying levels of missing data.

ConvLSTM performs well at lower missing steps (1 and 2) and achieves higher PSNR values than P-STMAE in these conditions.
As the number of missing steps increases, its performance deteriorates sharply, revealing sensitivity to temporal disruptions and difficulty in handling irregular sequences due to its end-to-end prediction in full-space.
ConvRAE exhibits a similar trend, with decent performance at lower missing steps but a noticeable decline as missing data increases.

The experiment demonstrates that P-STMAE maintains high performance across different missing ratios.
In contrast, the RNN-based models are more effective with limited missing data but struggle as gaps increase, highlighting the advantage of transformer-based attention mechanisms for handling irregular sampling.

\subsubsection{Nonlinear robustness analysis}\label{Xsec29-4.2.6}

We explore nonlinear robustness by introducing dilations into the sampling window with a fixed missing ratio of 0.5.
Dilation increases the time gaps between consecutive data points by a fixed factor, testing whether models can generalise when temporal dynamics become more irregular and chaotic.

Mathematically, given an original time series sequence:
\begin{align}
\mathbf{X} = \{\mathbf{x}_1, \mathbf{x}_2, \mathbf{x}_3, \dots, \mathbf{x}_T\},
\end{align}

where $\mathbf{x}_t$ represents the value at time step $t$, and a dilated sequence with dilation factor $d$ is defined as:
\begin{align}
\mathbf{X}_d = \{\mathbf{x}_1, \mathbf{x}_{1+d}, \mathbf{x}_{1+2d}, \dots, \mathbf{x}_{1+kd}\},
\end{align}

where $k$ is the largest integer satisfying $1 + kd \leq T$.
This means that every $d$-th element is sampled, increasing the time gap between consecutive points.

This operation simulates irregular, nonlinear time steps by introducing structured sparsity into the sequence.
The dilation parameter $d$ controls the degree of this gap expansion, allowing us to introduce more variability in the time structure.
As $d$ increases, the sequence becomes less regular and more nonlinear, challenging the models to generalise and capture complex, evolving temporal dynamics.

\begin{figure*}[!t]
\centering
\subfloat[Missing ratio analysis]{\includegraphics[width=\textwidth]{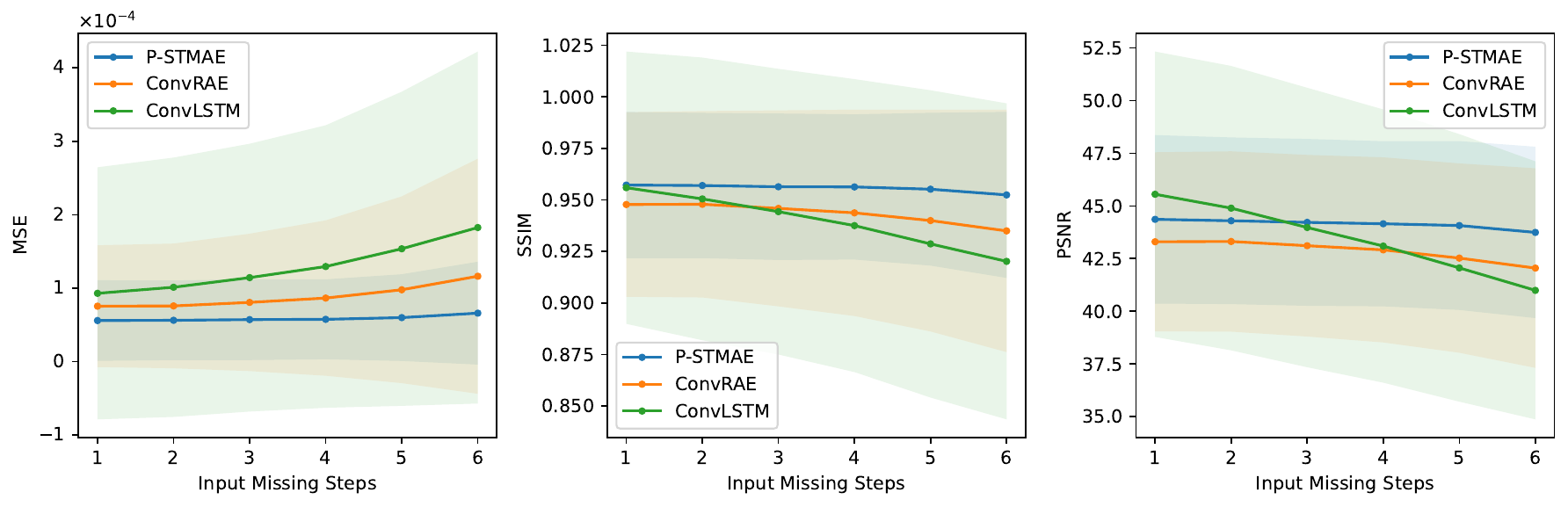}\label{fig:sw_missing_ratio}}
\\
\vspace{0.5cm}
\subfloat[Sampling dilation analysis]{\includegraphics[width=\textwidth]{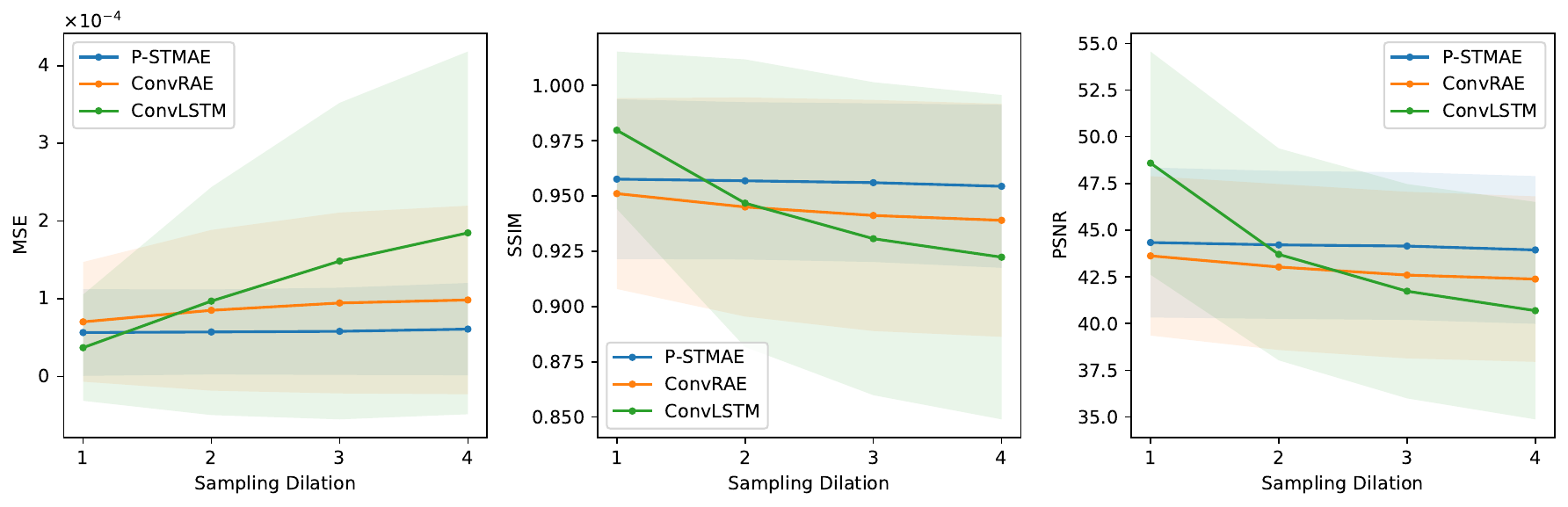}\label{fig:sw_dilations}}
\caption{Robustness analysis of P-STMAE on the shallow water dataset. \textbf{(a)} Performance comparison under varying numbers of missing steps in the input sequence with a length of 10. Each model is trained and evaluated with missing steps ranging from 1 to 6. P-STMAE demonstrates consistent performance and robustness, while the RNN-based models, especially ConvLSTM, show higher sensitivity to increasing missing steps. \textbf{(b)} Test performance comparison regarding the sampling dilations of data sequences. All models are separately trained on the shallow water dataset of different dilations.}
    \label{fig:sw_robustness}
\end{figure*}

Panel (b) of Fig.~\ref{fig:sw_robustness} shows the test performance results over different dilations.
The performance curves reveal distinct patterns: P-STMAE exhibits stable performance across different dilations, with MSE values remaining consistently low even as dilation increases, demonstrating robustness when faced with increasing nonlinearities.
In contrast, ConvLSTM shows sensitivity to dilation changes, performing well at minimal dilation but deteriorating rapidly as dilation increases, with error values rising.
ConvRAE shows moderate sensitivity, with performance declining gradually but remaining better than ConvLSTM at higher dilations.
This suggests that ConvLSTM struggles to generalise in complex scenarios where temporal dependencies become harder to capture due to high nonlinearities, likely because it operates directly in full physics space where temporal relations are chaotic on sparse features.

Both P-STMAE and ConvRAE show resilience to dilation, suggesting that latent-space architectures are better suited for handling chaotic data.
By processing spatial information into dense features via CAEs, they can manage complex, nonlinear patterns without being directly impacted by irregularities in the full physics space.
The performance gap between P-STMAE and ConvRAE further reveals the transformer-based capability in capturing irregular temporal dependencies compared to RNNs.

\subsection{Diffusion reaction test case}\label{Xsec30-4.3}

\subsubsection{Dataset description}\label{Xsec31-4.3.1}

The 2D diffusion-reaction equations are commonly employed to model phenomena where diffusion and reaction processes interact in a spatial domain, such as biological pattern formation.
Compared to the shallow water case, this case emphasises coupled nonlinear interactions and pattern formation.
The system consists of two nonlinearly coupled variables, the activator $u$ and the inhibitor $v$.
The equations governing their evolution are given by~\cite{takamoto2022pdebench}:
\begin{align}
    \frac{\partial u}{\partial t} &= \alpha_u \left(\frac{\partial^2\thinspace u}{\partial x^2} + \frac{\partial^2\thinspace u}{\partial y^2}\right) + F_u(u, v), \\
    \frac{\partial v}{\partial t} &= \alpha_v \left(\frac{\partial^2 v}{\partial x^2} + \frac{\partial^2 v}{\partial y^2}\right) + F_v(u, v).
\end{align}

Here, $\alpha_u$ and $\alpha_v$ denote the diffusion coefficients for the activator and inhibitor.
The reaction functions $F_u$ and $F_v$ follow the FitzHugh-Nagumo model:
\begin{align}
    F_u(u, v) &= u - u^3 - c - v, \\
    F_v(u, v) &= u - v,
\end{align}

where $c$ is a constant parameter that affects the reaction kinetics.
The domain for the simulation extends over $x, y \in [-1, 1]$ with time $t \in (0, 5]$.
The no-flow Neumann boundary conditions ensure that the flux of both $u$ and $v$ across the boundaries remains zero.

The training dataset, available from the PDEBench~\cite{takamoto2022pdebench} project, is discretised into $128 \times 128$ spatial grid points and 100 temporal steps, with 10,000 sample sequences.

The 2D diffusion-reaction dataset poses a challenge due to the nonlinear coupling between the activator and inhibitor, and its applicability to real-world problems such as biological pattern formation.

\begin{figure*}[!t]
\centerline{\includegraphics[width=\textwidth]{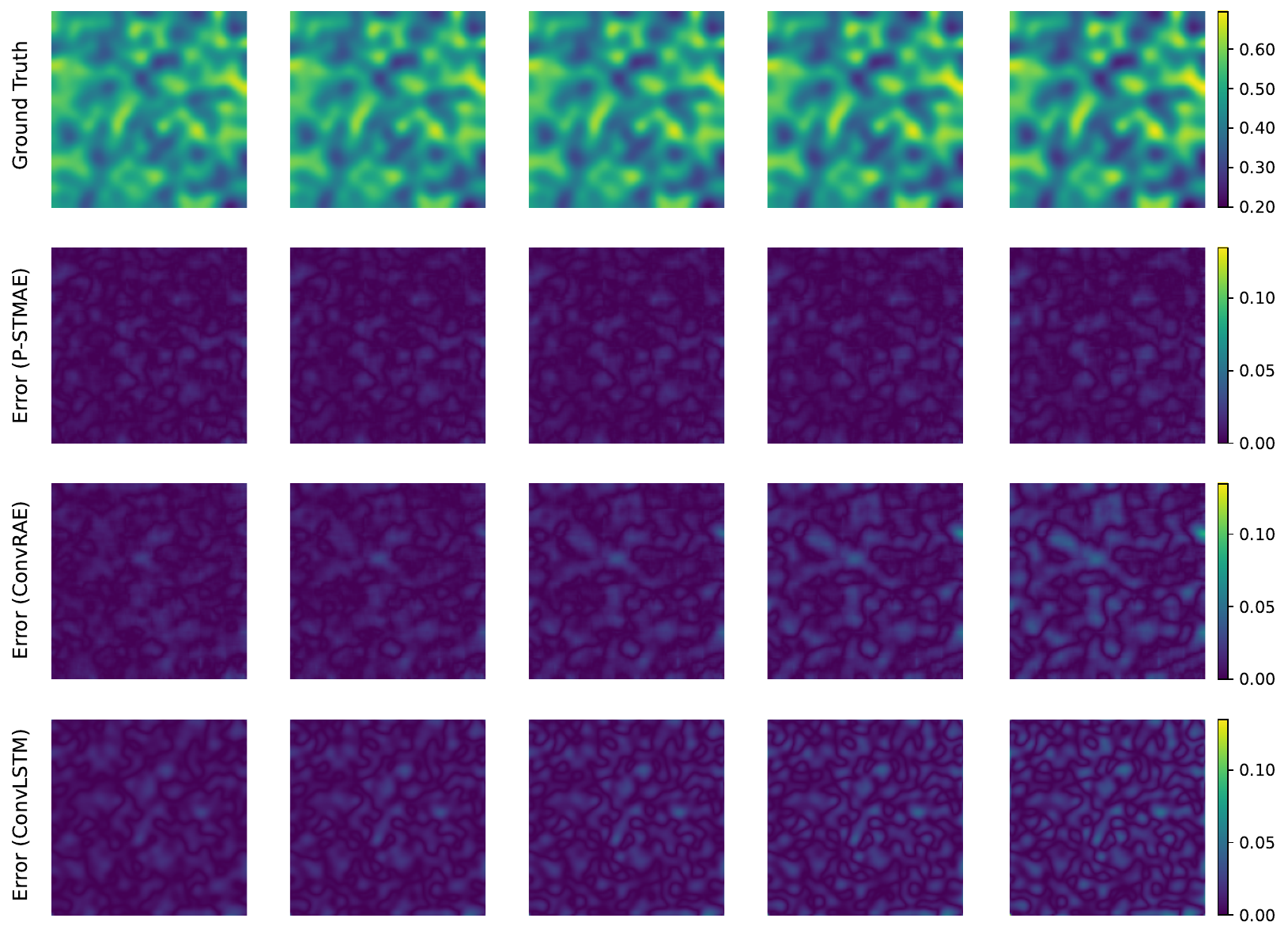}}
\caption{Ground truth (top) and error maps of P-STMAE, ConvRAE, and ConvLSTM for forecasting the variable $u$ in the diffusion-reaction dataset with a sampling dilation of 5. Columns represent successive forecasting steps. The results show that P-STMAE consistently achieves lower errors than the baselines, confirming its advantage.}
    \label{fig:dr_err_comparison}
\end{figure*}

\subsubsection{Model comparison}\label{Xsec32-4.3.2}

For the diffusion-reaction dataset (see Table~\ref{tab:combined_test_metrics}), P-STMAE achieves the lowest MSE among the baselines, indicating numerical accuracy in minimising pointwise error.
However, P-STMAE slightly underperforms in SSIM and PSNR compared to ConvLSTM, suggesting a trade-off between pixel-wise accuracy and higher-order spatial consistency (see Appendix Fig.~\ref{fig:dr_comparison}).
This may arise because ConvLSTM operates directly in the full physics space, potentially better preserving structural patterns and perceptual quality for complex coupled-variable systems, while P-STMAE's latent-space compression may introduce subtle spatial distortions despite pointwise accuracy.

Fig.~\ref{fig:dr_err_comparison} provides a visual comparison of prediction errors across all three models.
The error maps demonstrate that P-STMAE consistently achieves lower prediction errors than both baseline models across successive forecasting steps.
The spatial error patterns reveal that P-STMAE maintains higher accuracy in regions with complex pattern formations, where the activator and inhibitor variables exhibit strong coupling.
In contrast, ConvRAE and ConvLSTM show larger error magnitudes in areas where the reaction dynamics create intricate spatial structures (see Appendix Fig.~\ref{fig:dr_pred_err_comparison}).

\subsection{NOAA sea surface temperature test case}\label{Xsec33-4.4}

\subsubsection{Dataset description}\label{Xsec34-4.4.1}

The SST~\cite{huang2021improvements} dataset provides a long-term climate record of weekly sea surface temperature observations spanning the period from 1981 to 2018.
These data are collected from multiple sources, including satellites, ships, buoys, and Argo floats, and are then interpolated to produce a continuous global grid of temperature data.
The spatial resolution of the dataset is $360 \times 180$, corresponding to a global grid where each unit covers a 1-degree area of latitude and longitude.

The dataset consists of 1914 snapshots, each representing the global distribution of sea surface temperatures at weekly intervals.
These observations are stored as a single temperature variable.
The temporal and spatial continuity of the dataset makes it valuable for studying long-term climate trends, oceanic processes, and their influence on weather and marine ecosystems.

The SST dataset is widely used for analysing climate variability, detecting anomalies such as El Niño and La Niña, and studying oceanic heat content changes.
This dataset presents a complex spatiotemporal problem, as ocean temperatures are influenced by long-term climatic patterns, oceanic currents, and seasonal variations.

\begin{figure*}[!t]
\centerline{\includegraphics[width=\textwidth]{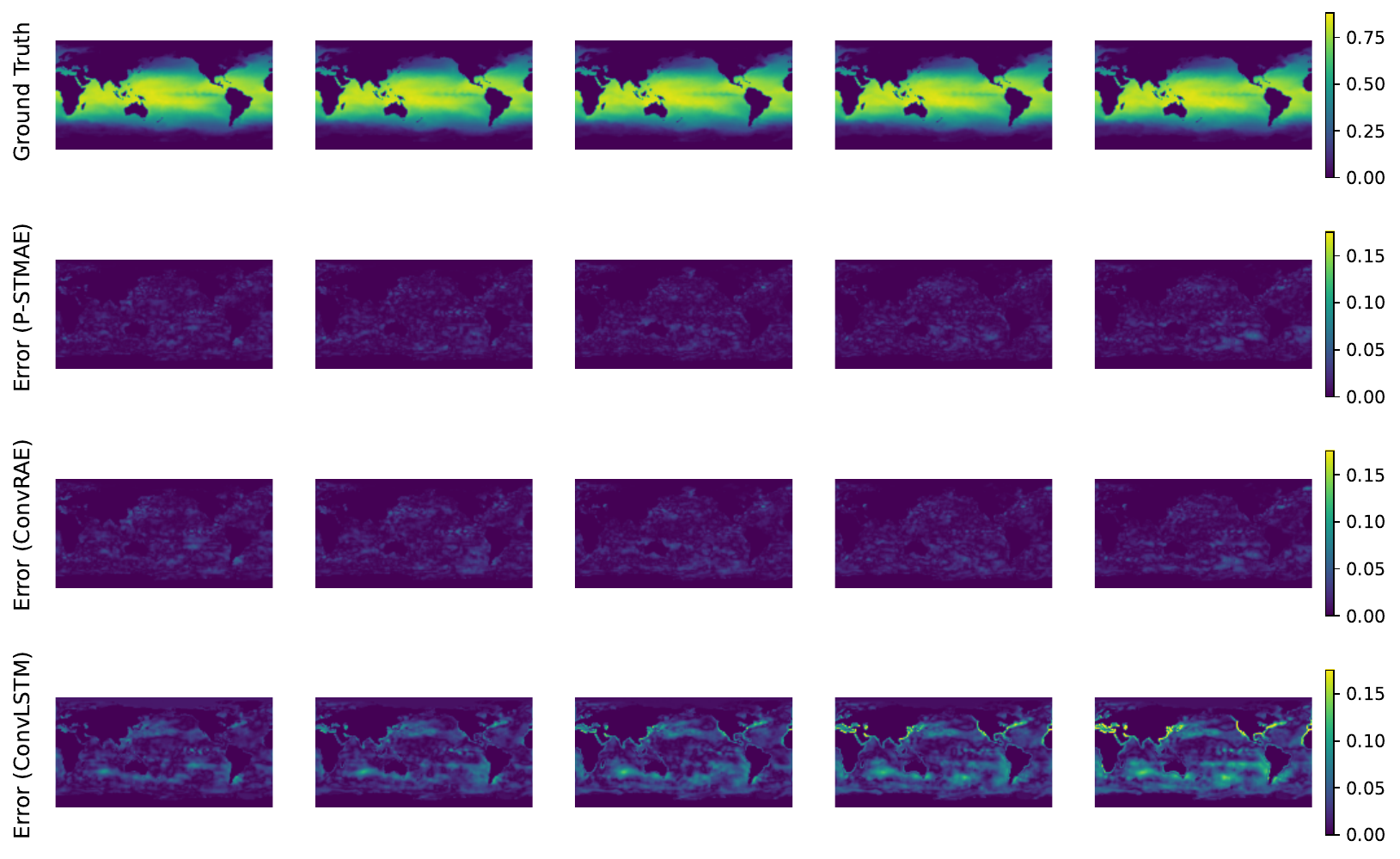}}
\caption{Error maps of P-STMAE, ConvRAE, and ConvLSTM predictions for the forecasting steps in the SST dataset.}
    \label{fig:sst_err_comparison}
\end{figure*}

\subsubsection{Model comparison}\label{Xsec35-4.4.2}

The performance on the SST dataset (Table~\ref{tab:combined_test_metrics}, missing ratio 0.5) demonstrates P-STMAE's strong generalisation to real-world climate data.
P-STMAE delivers the strongest overall performance, achieving the lowest MSE ($8.02 \times 10^{-5}$), highest SSIM (0.9817), and highest PSNR (41.03) compared with the baselines.
The transformer architecture enables P-STMAE to better handle irregularities and capture long-term dependencies in the SST dataset, leading to more accurate temporal predictions.
The performance of ConvRAE closely follows P-STMAE, while ConvLSTM falls behind, suggesting that latent-space compression is crucial for modelling real-world systems, and reduces the impact of noise and chaotic patterns that ConvLSTM struggles to capture at global scale.

Fig.~\ref{fig:sst_err_comparison} presents error maps comparing all three models across forecasting steps.
The global error patterns reveal that P-STMAE achieves the smallest prediction errors across most oceanic regions, with strong performance in areas with complex temperature gradients such as ocean fronts and upwelling zones. 
The error distribution shows that P-STMAE maintains consistent accuracy across different latitudinal bands and oceanic basins, demonstrating its ability to capture both large-scale climate patterns and regional temperature variations.
In contrast, ConvLSTM exhibits larger and more spatially widespread errors in regions with strong seasonal variations and complex current systems, highlighting the challenges of full-space modelling for global-scale climate data (see Appendix Fig.~\ref{fig:sst_pred_err_comparison}).

\section{Conclusion and future work}\label{Xsec36-5}
\label{ch:conclusion}

In this paper, we introduced the P-STMAE, a novel model designed to address irregular time series prediction in high-dimensional dynamical systems.
By integrating CAEs with transformer-based masked autoencoders, P-STMAE employs placeholder-based attention to handle missing data and irregular time steps directly without preprocessing Table~\ref{tab:lambda_ablation}.
Experiments across synthetic PDE benchmarks and real-world SST data demonstrated its robustness, computational efficiency, and better accuracy compared to traditional RNN-based approaches Fig.~\ref{fig:sst_pred_err_comparison}.

A key advantage of P-STMAE is its latent-space masked training, which enables efficient processing of high-dimensional data.
By operating in a compressed representation, the model reduces computational cost while maintaining spatiotemporal pattern learning.
This efficiency is beneficial for large-scale systems where computational resources are constrained.

Despite its promising performance, several limitations highlight areas for future research.
The quadratic complexity of the transformer's global self-attention poses challenges for processing very long sequences.
Exploring alternatives like local or sparse attention mechanisms could enhance scalability.
To efficiently handle relative time embedding in irregular time series, future work could consider advanced positional embedding techniques, such as Attention with Linear Biases (ALiBi)~\cite{press2021alibi} and Rotary Position Embedding (RoPE)~\cite{su2021roformer}, which can better capture relative temporal relationships without explicit positional encodings.
Additionally, the reliance on the convolutional autoencoder may introduce a bottleneck, potentially limiting reconstruction fidelity.
Future work could investigate advanced physical field encoding techniques, such as VAEs or Vision Transformers, to overcome this limitation.
Furthermore, the observed trade-off between minimizing point-wise prediction errors and preserving structural fidelity in spatiotemporal data remains an open challenge.
Future research could focus on multi-objective optimisation strategies that balance numerical accuracy with the preservation of global structures.
Finally, while the encouraging results on SST provide an initial demonstration of real-world applicability, broader validation on diverse real-world datasets will be essential to fully establish the model's generalizability.
In summary, this work advances irregular time series forecasting for high-dimensional dynamical systems.
P-STMAE offers a purely data-driven, adaptable, and computationally efficient solution, positioning it as a promising tool for scientific and industrial applications requiring accurate prediction of complex spatiotemporal systems.

\section*{CRediT authorship contribution statement}
\textbf{Kewei Zhu:} Writing -- original draft, Software, Investigation, Formal analysis, Data curation.
\textbf{Yanze Xin:} Writing -- original draft, Methodology, Investigation, Formal analysis, Data curation.
\textbf{Jinwei Hu:} Writing -- review \& editing, Validation, Methodology.
\textbf{Xiaoyuan Cheng:} Writing -- review \& editing, Validation, Investigation.
\textbf{Yiming Yang:} Writing -- review \& editing, Validation, Investigation.
\textbf{Sibo Cheng:} Writing -- review \& editing, Validation, Investigation, Funding acquisition, Conceptualization.

\section*{Data availability}
The code of this study is available at \url{https://github.com/RyanXinOne/PSTMAE}.

\section*{Declaration of competing interest}
 The authors declare that they have no known competing financial
interests or personal relationships that could have appeared to
influence the work reported in this paper.

\appendix
\counterwithin{figure}{section}
\counterwithin{table}{section}
\renewcommand{\thefigure}{\thesection\arabic{figure}}
\renewcommand{\thetable}{\thesection\arabic{table}}

\section{Detailed Visualization of Model Predictions}

\subsection{Model architecture}\label{Xsec37-A.1}

\begin{table}[h]
    \caption{Network structures of CAE encoder (left) and decoder (right) used in the P-STMAE. All of the convolutions and transpose convolutions use the kernel size of $3 \times 3$ and the same padding. The input dimension and channel number can vary depending on the dataset.}
\label{tab:cae_structure}
        \begin{tabular}{lll}
\toprule
            {Layer Type} & {Output Shape} & {Activation} \\
            \midrule
            Input               & (128, 128, 3)         &                     \\
            Conv 2D             & (128, 128, 8)         & GELU                \\
            Conv 2D             & (64, 64, 16)          & GELU                \\
            Conv 2D             & (32, 32, 32)          & GELU                \\
            Conv 2D             & (16, 16, 64)          & GELU                \\
            Conv 2D             & (8, 8, 128)           & GELU                \\
            Linear              & (128)                 &                     \\
        \midrule
            Input               & (128)                 &                     \\
            Linear              & (8, 8, 128)           & GELU                \\
            TransConv 2D        & (16, 16, 64)          & GELU                \\
            TransConv 2D        & (32, 32, 32)          & GELU                \\
            TransConv 2D        & (64, 64, 16)          & GELU                \\
            TransConv 2D        & (128, 128, 8)         & GELU                \\
            Conv 2D             & (128, 128, 3)         & Sigmoid             \\
        \bottomrule
\end{tabular}
\end{table}

\section{Shallow Water Test Case}

\begin{table}[h]
\caption{Ranges of computational parameters used in the shallow water equation simulations. All parameters are uniformly sampled.}
\begin{tabular}{llll}
\toprule
{Parameter} & {Symbol} & {Min Value} & {Max Value} \\
\midrule
Initial bump centre (x) & $p_x$ & 54.00 & 74.00 \\
Initial bump centre (y) & $p_y$ & 54.00 & 74.00 \\
Bump height & $h$ & 0.05 & 0.20 \\
Bump radius & $r$ & 8.94 & 12.65 \\
Friction coefficient & $b$ & 0.02 & 2.00 \\
Snapshot interval (steps) & -- & 60.00 & 100.00 \\
\bottomrule
\end{tabular}
\label{tab:sw_parameters}
\end{table}

\begin{table}[h]
\caption{Quantitative ablation study on $\lambda$.}
\label{tab:lambda_ablation}
\begin{tabular}{llll}
\toprule
$\lambda$ & MSE & SSIM & PSNR \\
\midrule
0.01 & $5.85 \times 10^{-5}$ & 0.9276 & 41.02 \\
0.02 & $5.25 \times 10^{-5}$ & 0.9317 & 41.52 \\
0.05 & $4.72 \times 10^{-5}$ & 0.9264 & 40.82 \\
0.10 & $5.20 \times 10^{-5}$ & 0.9315 & 41.60 \\
0.20 & $6.10 \times 10^{-5}$ & 0.9177 & 39.56 \\
0.50 & $8.00 \times 10^{-5}$ & 0.9356 & 42.07 \\
0.60 & $2.17 \times 10^{-4}$ & 0.9037 & 35.46 \\
0.70 & $2.20 \times 10^{-4}$ & 0.9017 & 35.38 \\
1.00 & $2.18 \times 10^{-4}$ & 0.9037 & 35.43 \\
\bottomrule
\end{tabular}

\end{table}

\begin{figure*}[H]
\centerline{\includegraphics[width=\textwidth]{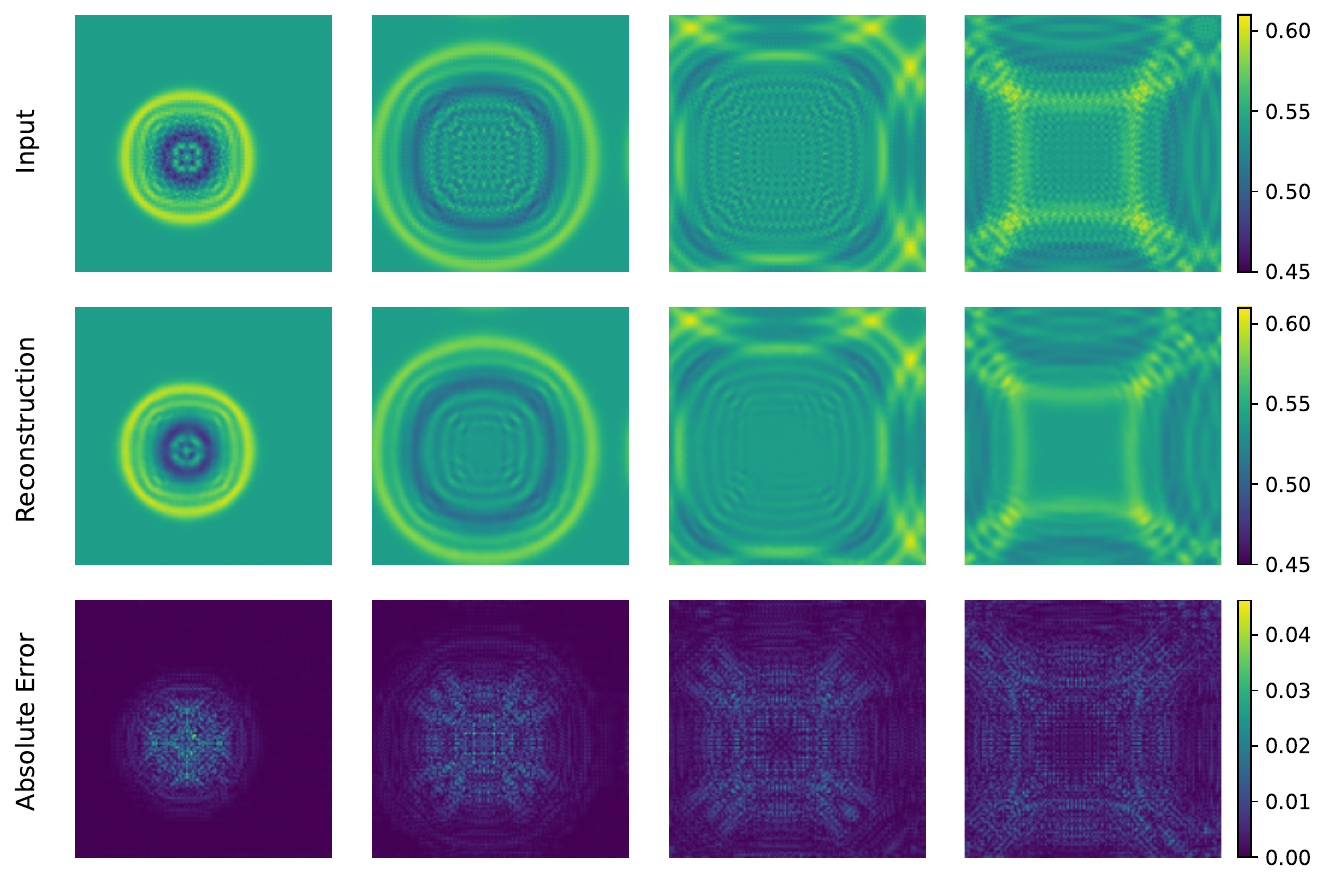}}
\caption{Visualisation of the trained CAE model performance on the variable $h$ of shallow water samples.}
    \label{fig:sw_cae_evaluation}
\end{figure*}

\begin{figure*}[H]
\centerline{\includegraphics[width=\textwidth]{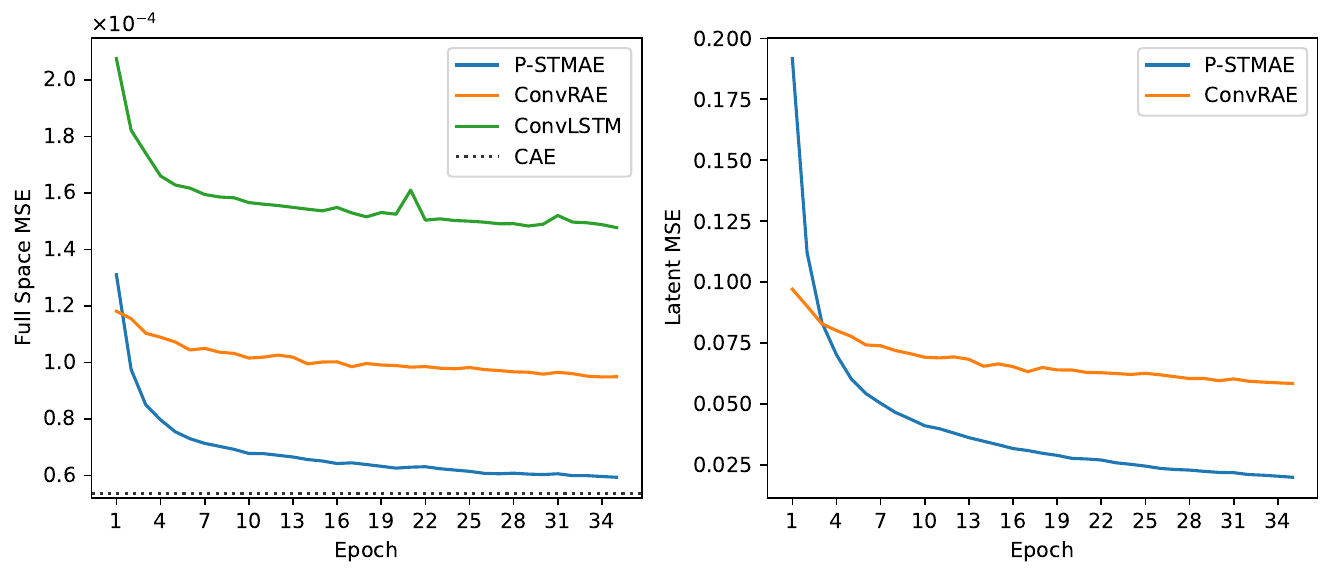}}
\caption{Validation MSEs on the shallow water dataset with the sampling dilation of 3. \textbf{Left:} Full space MSE curves of the P-STMAE, ConvRAE, and ConvLSTM. The dotted line shows the performance of the trained CAE. \textbf{Right:} Latent space MSE curves of the P-STMAE and ConvRAE. Note that the ConvLSTM does not use latent representations.}
    \label{fig:sw_comparison}
\end{figure*}

\begin{figure*}[H]
\centerline{\includegraphics[width=\textwidth]{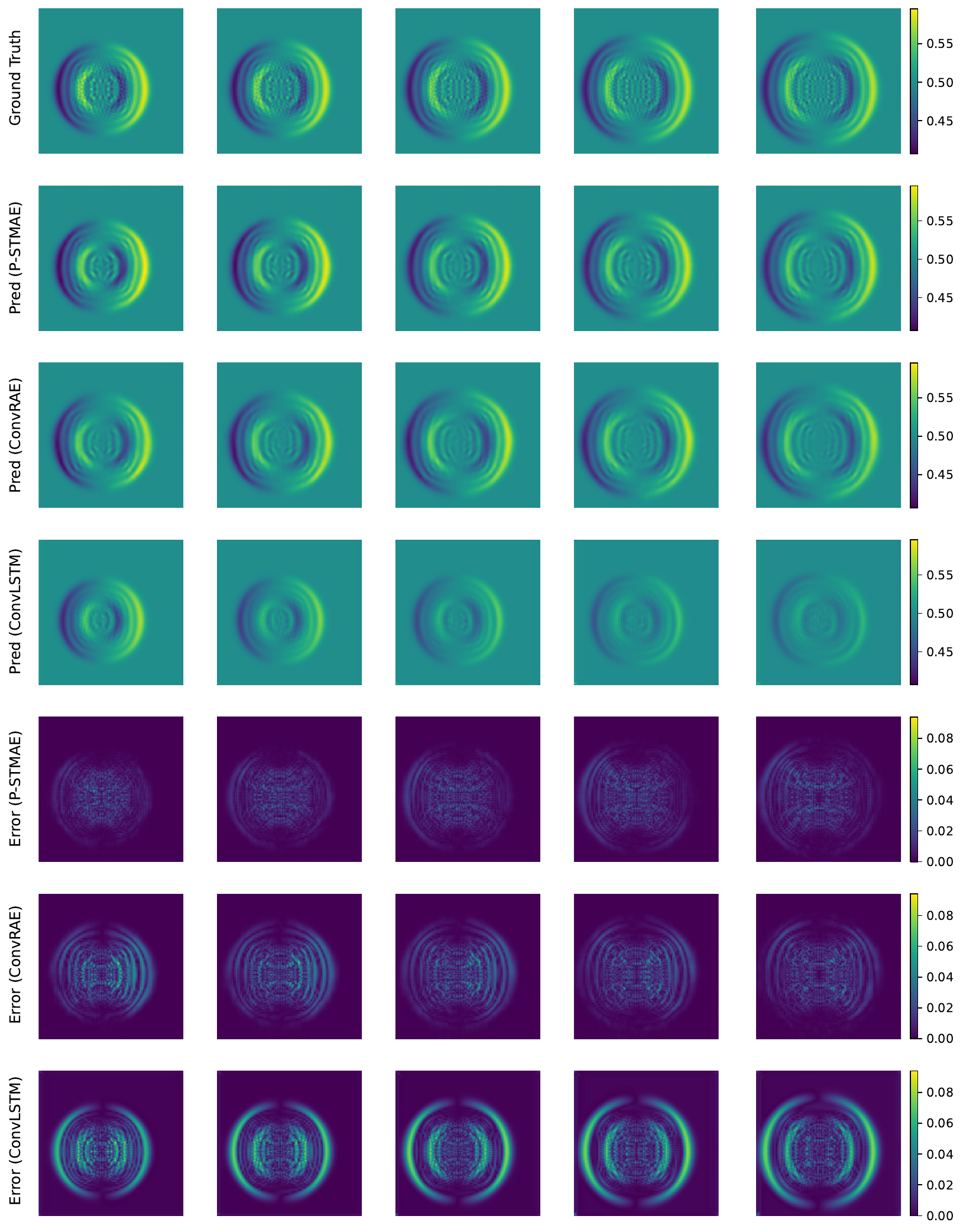}}
\caption{Model predictions and error maps for the forecasting steps of the variable $u$ in the shallow water dataset with a sampling dilation of 3. The first row shows the ground truth input, rows 2--4 show the model-predicted outputs of the three models, and rows 5--7 show the pixel-level absolute differences between the two.}
    \label{fig:sw_pred_err_comparison}
\end{figure*}

\begin{figure*}[H]
\centerline{\includegraphics[width=\textwidth]{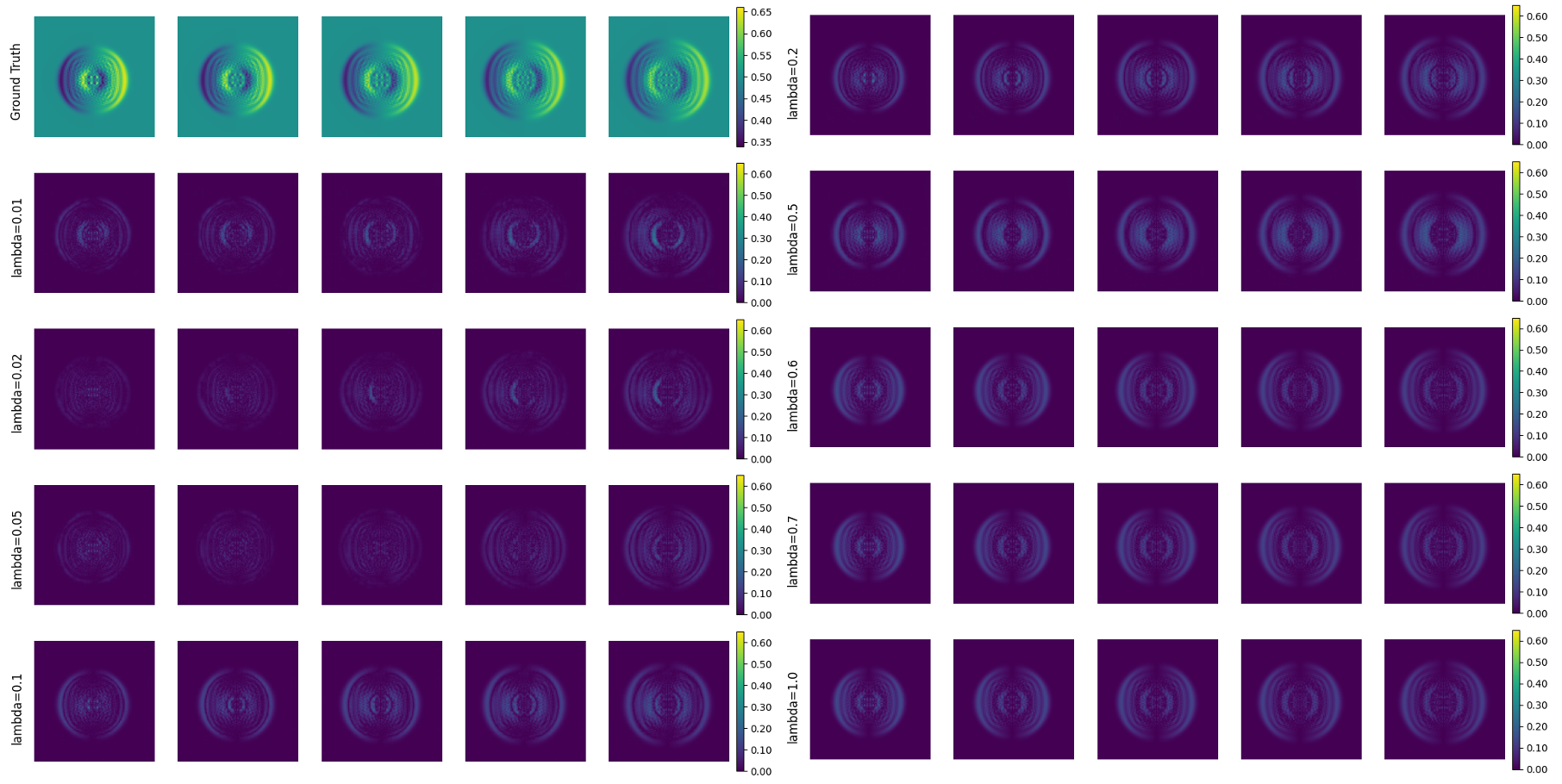}}
\caption{Model error maps for different values of $\lambda$ across forecasting steps in the shallow water dataset.}
    \label{fig:lambda_ablation}
\end{figure*}

\section{Diffusion Reaction Test Case}

\begin{figure}[H]
\centerline{\includegraphics[width=\textwidth]{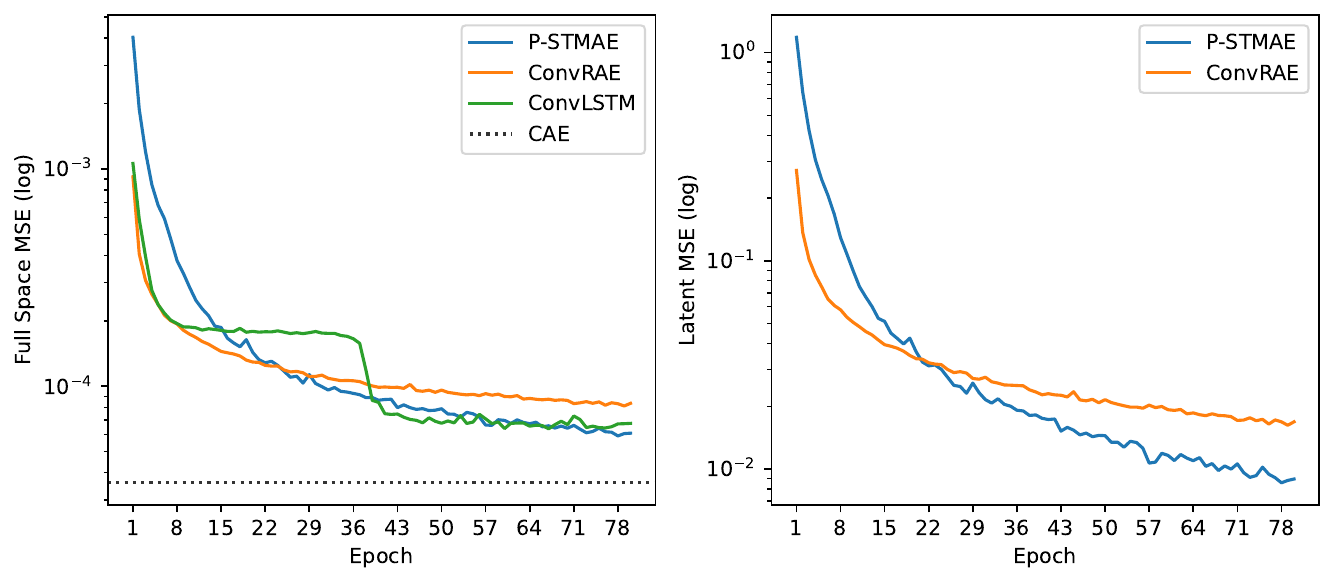}}
\caption{Validation MSEs on the diffusion reaction dataset with the sampling dilation of 5.}
    \label{fig:dr_comparison}
\end{figure}

\begin{figure*}[H]
\centerline{\includegraphics[width=\textwidth]{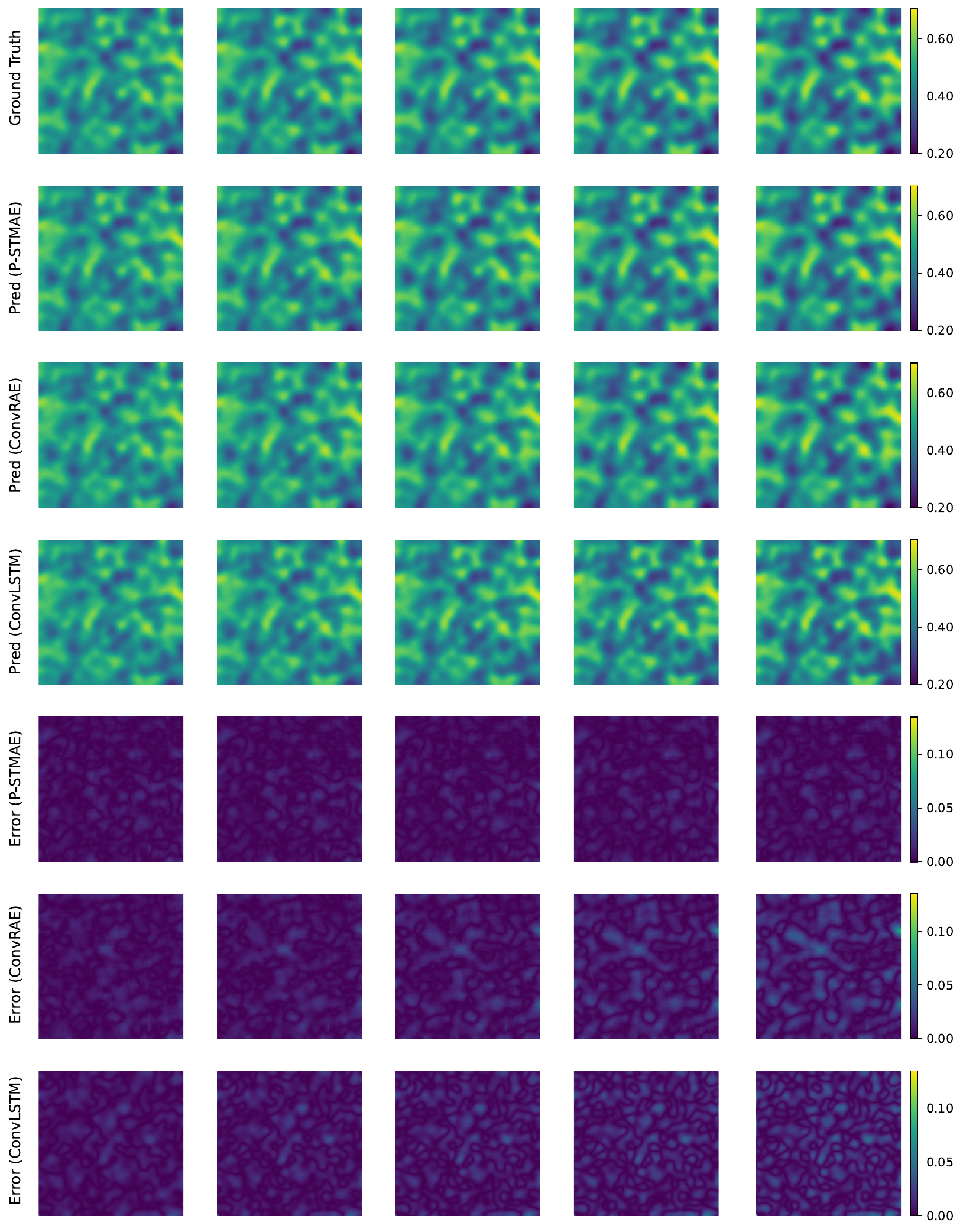}}
\caption{Model predictions and error maps for the forecasting steps of the variable $u$ in the diffusion reaction dataset with a sampling dilation of 5.}
    \label{fig:dr_pred_err_comparison}
\end{figure*}

\section{NOAA Sea Surface Temperature Test Case}

\begin{figure}[H]
\centerline{\includegraphics[width=\textwidth]{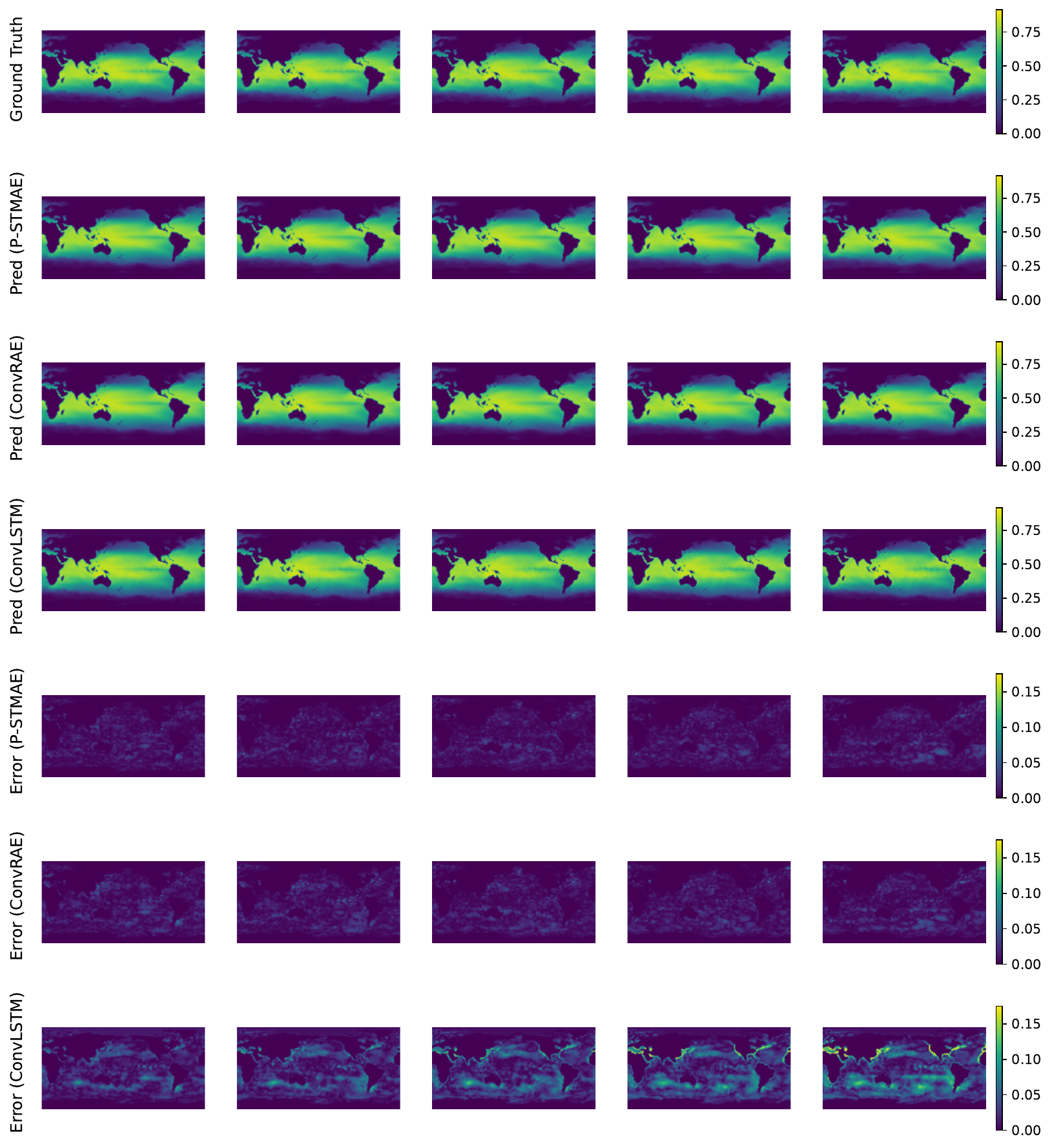}}
\caption{Model predictions and error maps for the forecasting steps in the SST dataset.}
    \label{fig:sst_pred_err_comparison}
\end{figure}

\end{document}